\documentclass[10pt,twocolumn,letterpaper]{article}

\usepackage{cvpr}
\usepackage{times}
\usepackage{epsfig}
\usepackage{graphicx}
\usepackage{amsmath}
\usepackage{amssymb}
\usepackage{tabularx}
\usepackage{multirow}
\usepackage{float}%
\usepackage[labelsep=period]{caption}
\usepackage{subcaption} %
\usepackage[export]{adjustbox} %
\usepackage[para]{threeparttable} %
\usepackage{fancyhdr} %
\usepackage{booktabs} %
\usepackage[pagebackref=true,breaklinks=true,letterpaper=true,colorlinks,bookmarks=false]{hyperref}

\cvprfinalcopy %

\def\cvprPaperID{119} %

\ifcvprfinal\fi
\begin{document}

\newcommand{\todo}[1]{{\color{blue} #1}}    %

\title{Proposal Flow}

\author{Bumsub Ham\textsuperscript{1,}\thanks{indicates equal contribution}~~\textsuperscript{,}\thanks{WILLOW project-team, D\'epartement d'Informatique de l'Ecole Normale
Sup\'erieure, ENS/Inria/CNRS UMR 8548.} \quad\quad\quad Minsu Cho\textsuperscript{1,}\footnotemark[1]~~\textsuperscript{,}\footnotemark[2] \quad\quad\quad Cordelia Schmid\textsuperscript{1,}\thanks{Thoth project-team, Inria Grenoble Rh\^one-Alpes, Laboratoire Jean Kuntzmann.} \quad\quad\quad Jean Ponce\textsuperscript{2,}\footnotemark[2] \vspace*{0.2cm}\\
{\textsuperscript{1}Inria \quad\quad \textsuperscript{2}\'{E}cole Normale Sup\'erieure / PSL Research University}}

\maketitle
\thispagestyle{empty}

\begin{abstract}
\vspace{-0.4cm}
Finding image correspondences remains a challenging problem in the
presence of intra-class variations and large changes in scene layout.~Semantic flow methods are designed to handle images depicting different instances of the same object or scene category. 
We introduce a novel approach to semantic flow, dubbed proposal flow, that establishes reliable correspondences using object proposals. 
Unlike prevailing semantic flow
approaches that operate on pixels or regularly sampled local regions,
proposal flow benefits from the characteristics of modern object
proposals, that exhibit high repeatability at multiple scales, and can
take advantage of both local and geometric consistency constraints
among proposals. We also show that proposal flow can effectively be
transformed into a conventional dense flow field. We introduce a new
dataset that can be used to evaluate both general semantic flow
techniques and region-based approaches such as proposal flow. We use
this benchmark to compare different matching algorithms, object
proposals, and region features within proposal flow, to the state of
the art in semantic flow. This comparison, along with experiments on
standard datasets, demonstrates that proposal flow significantly
outperforms existing semantic flow methods in various settings.
\end{abstract}
\vspace{-0.5cm}
\section{Introduction}
\vspace{-0.2cm}
Classical approaches to finding correspondences across images are designed to handle scenes that contain the same objects with moderate view point variations in applications such as stereo matching~\cite{okutomi1993multiple,rhemann2011fast}, optical flow~\cite{horn1993determining,weinzaepfel2015deepmatching,weinzaepfel2013deepflow}, and wide-baseline matching~\cite{matas2004robust,yang2014daisy}. {\em Semantic flow} methods, such as SIFT Flow~\cite{liu2011sift} for example, on the other hand, are designed to handle a much higher degree of variability in appearance and scene layout, typical of images depicting different instances of the same object or scene category.~They have proven useful for many tasks such as scene recognition, image registration, semantic segmentation, and image editing and synthesis~\cite{hacohen2011non,kim2013deformable,liu2011sift,yang2014daisy,zhou2015flowweb}. 
In this context, however, appearance and shape variations may confuse similarity measures for local region matching, and prohibit the use of strong geometric constraints (e.g., epipolar geometry, limited disparity range). Existing approaches to semantic flow are thus easily distracted by scene elements specific to individual objects and image-specific details (e.g., background, texture, occlusion, clutter). This is the motivation for our work, where we use robust region correspondences to focus on regions containing prominent objects and scene elements rather than clutter and distracting details.

\begin{figure}[t]
\captionsetup{font={small}}
\centering
\includegraphics[width=\linewidth]{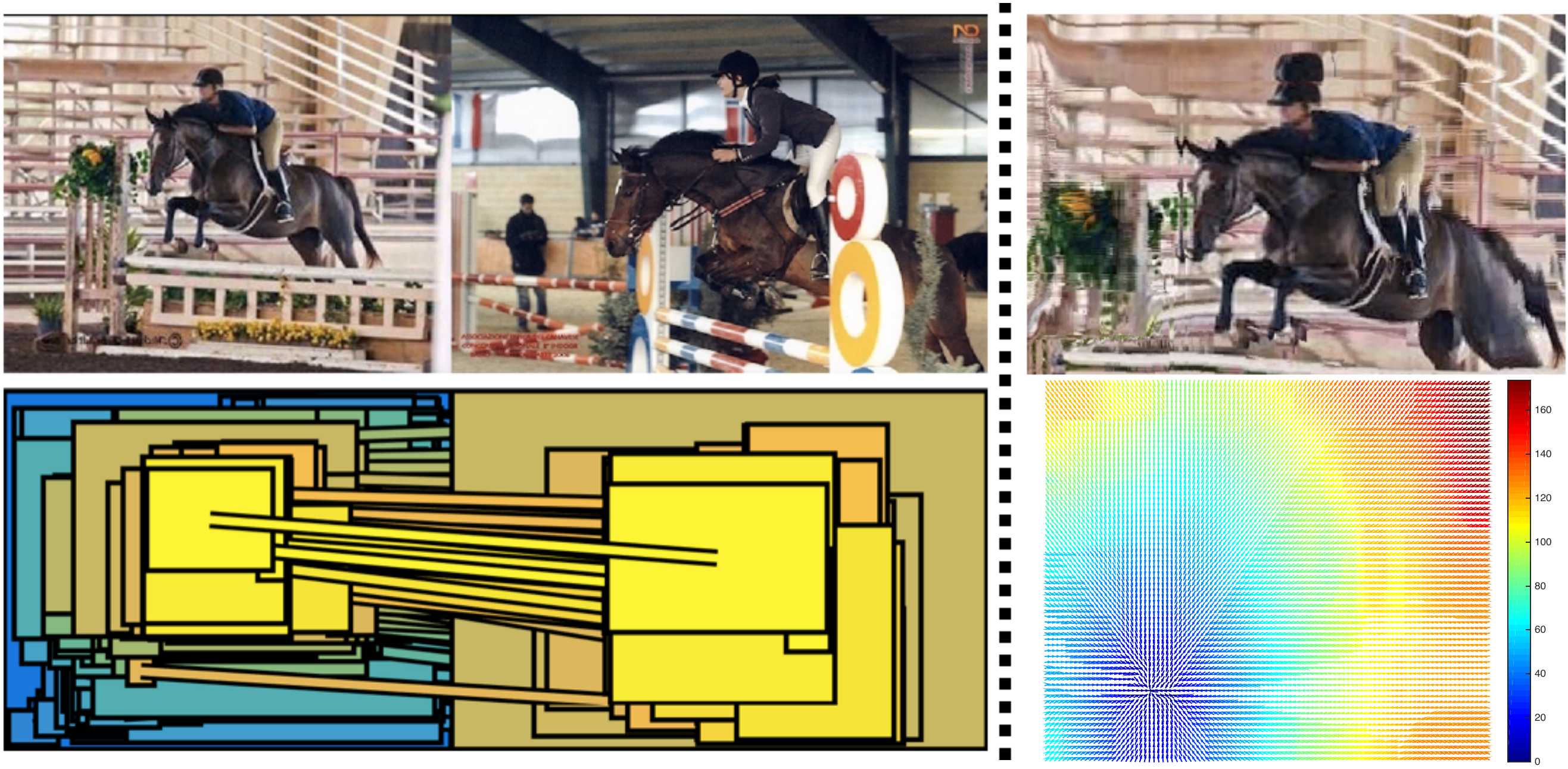}
\vspace{-0.1cm}
\begin{minipage}{0.625\linewidth}
\centering
\small{(a) Region-based semantic flow.}
\end{minipage}
\vspace{-0.1cm}
\begin{minipage}{0.365\linewidth}
\centering
\small{(b) Dense flow field.}
\end{minipage}

\caption{Proposal flow generates a reliable semantic flow between similar images using local and geometric consistency constraints among object proposals, and it can be transformed into a dense flow field. (a) Region-based semantic flow.~(b) Dense flow field and image warping using the flow field. \bf{(Best viewed in color.)}}
\label{fig:teaser}	
\end{figure}

Concretely, we introduce an approach to semantic flow
computation, called {\em proposal flow}, that establishes region
correspondences using object proposals and their geometric relations
(Fig.~\ref{fig:teaser}). Unlike previous semantic flow
algorithms~\cite{BristowVL15,hacohen2011non,hassner2012sifts,hur2015generalized,kim2013deformable,liu2011sift,qiu2014scale,tau2014dense,trulls2013dense,yang2014daisy,zhou2015flowweb},
that use regular grid structures for local region generation and
matching, we leverage a large number of multi-scale object
proposals~\cite{arbelaez2014multiscale,hosang2015what,manen2013prime,uijlings2013selective,zitnick2014edge},
as now widely used in object
detection~\cite{girshickICCV15fastrcnn,kaiming14ECCV}. The proposed
approach establishes region correspondences by exploiting their visual
features and geometric relations in an efficient manner, and generates
a region-based semantic flow composed of object proposal matches.~We
also show that the proposal flow can be effectively transformed into a
conventional dense flow field. Finally, we introduce a new dataset
that can be used to evaluate both general semantic flow techniques and
region-based approaches such as proposal flow. We use this benchmark
to compare different matching algorithms, object proposals, and region
features within proposal flow, to the state of the art in semantic
flow. This comparison, along with experiments on standard datasets,
demonstrates that proposal flow significantly outperforms existing
semantic flow methods in various settings.

\vspace{-0.2cm}
\section{Related work}
\vspace{-0.2cm}
Correspondence problems involve a broad range of topics beyond the scope of this paper. Here we briefly describe the context of our approach, and only review representative works pertinent for ours. 
Classical approaches to stereo matching and optical flow estimate pixel-level dense correspondences between two nearby images of the same scene~\cite{horn1993determining,matas2004robust,okutomi1993multiple}. 
While advances in invariant feature detection and description have revolutionized object recognition and reconstruction in the past 15 years, research on image matching and alignment between images have long been dominated by instance matching with the same scene and objects~\cite{forsyth2011modern}. Unlike these, several recent approaches to semantic flow focus on handling images containing different scenes and objects.~Graph-based matching algorithms~\cite{cho2012progressive,duchenne2011graph} attempt to find category-level feature matches by leveraging a flexible graph representation of images, but they commonly handle sparsely sampled or detected features due to their computational complexity. Inspired by classic optical flow algorithms, Liu~\etal pioneered the idea of dense correspondences across different scenes, and proposed the SIFT Flow~\cite{liu2011sift} algorithm that uses a multi-resolution image pyramid together with a hierarchical optimization technique for efficiency. Kim~\etal~\cite{kim2013deformable} extended the approach by inducing a multi-scale regularization with a hierarchically connected pyramid of grid graphs. More recently,~Long~\etal~\cite{long2014convnets} have investigated the effect of pretrained ConvNet features on the SIFT Flow algorithm, and Bristow~\etal~\cite{BristowVL15} have proposed an exemplar-LDA approach that improves the performance of semantic flow. Despite differences in graph construction, optimization, and similarity computation, existing semantic flow approaches share grid-based regular sampling and spatial regularization: The appearance similarity is defined at each region or pixel on (a pyramid of)  regular grids, and spatial regularization is imposed between neighboring regions in the pyramid models~\cite{kim2013deformable,liu2011sift}. In contrast, our work builds on generic object proposals with diverse spatial supports~\cite{arbelaez2014multiscale,hosang2015what,manen2013prime,uijlings2013selective,zitnick2014edge}, and uses an irregular form of spatial regularization based on co-occurrence and overlap of the proposals. We show that the use of local regularization with object proposals yields substantial gains in generic region matching and semantic flow, in particular when handling images with significant clutter and intra-class variations. 

Object proposals~\cite{arbelaez2014multiscale,hosang2015what,manen2013prime,uijlings2013selective,zitnick2014edge} have originally been developed  for object detection, where they are used to reduce the search space as well as false alarms. They are now an important component in many state-of-the-art detection pipelines~\cite{girshickICCV15fastrcnn,kaiming14ECCV}. Despite their success on object detection and segmentation, they have seldom been used in matching tasks~\cite{cho2015unsupervised,jiang2015matching}. In particular, while Cho \etal~\cite{cho2015unsupervised} have shown that object proposals are useful for region matching due to their high repeatability on salient part regions, the use of object proposals has never been thoroughly investigated in semantic flow computation. The approach proposed in this paper is a first step in this direction, and we explore how the choice of object proposals, matching algorithms, and features affects matching robustness and accuracy. %

\vspace{-0.4cm}
\paragraph{Contributions.} The contributions of this paper are three-fold: (i) 
We introduce the proposal flow approach to establishing robust region
correspondences between related, but not identical scenes using object
proposals. (ii) We introduce a benchmark for semantic flow that can be
used to evaluate both general semantic flow algorithms and region
matching methods.~(iii) We demonstrate the
advantage of proposal flow over state-of-the-art semantic
flow methods through extensive experimental evaluations.
\vspace{-0.1cm}
\section{Proposal flow} 
\vspace{-0.2cm}
Proposal flow can use any type of object
proposals~\cite{arbelaez2014multiscale,hosang2015what,manen2013prime,uijlings2013selective,zitnick2014edge}
as candidate regions for matching two images of related scenes. In
this section, we introduce a probabilistic model for region matching,
and describe three matching strategies including two baselines and a 
new one using local regularization. We then describe our approach
to generating a dense flow field from the region matches.
\begin{figure*}
\captionsetup{font={small}}
\captionsetup[subfigure]{aboveskip=-0.5pt,belowskip=-0.5pt}
\captionsetup{belowskip=0pt}
\centering
	\begin{subfigure}{0.326\textwidth}
	\includegraphics[width=\textwidth, frame]{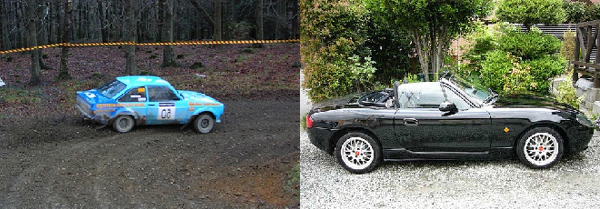}
	\caption{Input images.}
	\end{subfigure}
	\begin{subfigure}{0.326\textwidth}
	\includegraphics[width=\textwidth, frame]{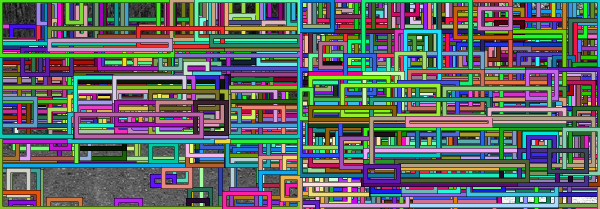}
	\caption{Object proposals~\cite{uijlings2013selective}.}
	\end{subfigure}
	\begin{subfigure}{0.326\textwidth}
	\includegraphics[width=\textwidth, frame]{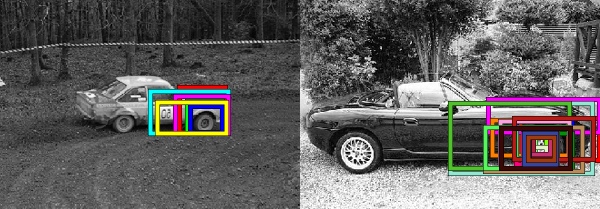}
	\caption{Object proposals near the front wheel.}
	\end{subfigure}

	\begin{subfigure}{0.326\textwidth}
	\includegraphics[width=\textwidth, frame]{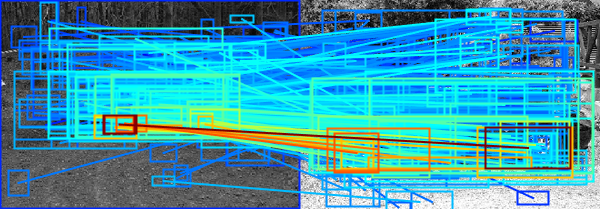}
	\caption{NAM.}
	\end{subfigure}
	\begin{subfigure}{0.326\textwidth}
	\includegraphics[width=\textwidth, frame]{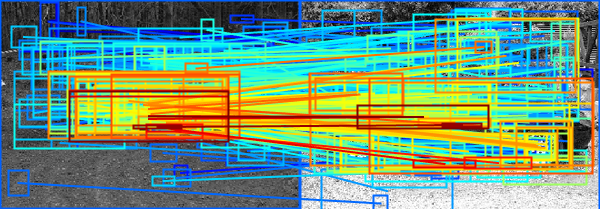}
	\caption{PHM~\cite{cho2015unsupervised}.}
	\end{subfigure}
	\begin{subfigure}{0.326\textwidth}
	\includegraphics[width=\textwidth, frame]{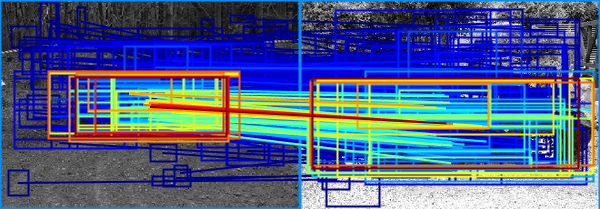}
	\caption{LOM.}
	\end{subfigure}
	\vfill
	\vspace{-0.2cm}
\caption{{\bf Top:} (a-b) Two images and their object
  proposals~\cite{uijlings2013selective}.  (c) Multi-scale object
  proposals contain the same object or parts, but they are not
  perfectly repeatable across different images. {\bf Bottom:} In
  contrast to NAM (d), PHM~\cite{cho2015unsupervised} (e) and LOM (f)
  both exploit geometric consistency, which regularizes proposal
  flow. In particular, LOM imposes local smoothness on offsets between
  neighboring regions, avoiding the problem of using a global
  consensus on the offset in PHM~\cite{cho2015unsupervised}. The
  matching score is color-coded for each match (red: high, blue:
  low). The HOG descriptor~\cite{dalal2005histograms} is used for
  appearance matching in this example. \bf{(Best viewed in color.)}}
\label{fig:compare_match}	%
\end{figure*} 
\vspace{-0.1cm}
\subsection{A Bayesian model for region matching}
\vspace{-0.1cm}
Let us suppose that two sets of object proposals $\mathcal{R}$ and $\mathcal{R}'$ have been extracted from images $\mathcal{I}$ and $\mathcal{I}'$ (Fig.~\ref{fig:compare_match}(a-b)). A proposal $r$ in $\mathcal{R}$ is an image region $r=(f,s)$ with appearance feature $f$ and spatial support $s$.
The appearance feature represents a visual descriptor for the region (e.g., SPM~\cite{lazebnik2006beyond} , HOG~\cite{dalal2005histograms}, ConvNet~\cite{krizhevsky2012imagenet}), and the spatial support describes the set of all pixel positions in the region, that forms a rectangular box in this work. Given the data $\mathcal{D} = (\mathcal{R}, \mathcal{R}')$, we wish to estimate a posterior probability of the event $r \mapsto r'$ meaning that proposal $r$ in $\mathcal{R}$ matches proposal $r'$ in $\mathcal{R}'$:\vspace{-0.1cm}  
\begin{equation}
\label{eq:similairty}
	p(r \mapsto r' \mid \mathcal{D}) = p(f \mapsto f' ) p(s \mapsto s' \mid \mathcal{D}), \vspace{-0.1cm}  
\end{equation}
where we decouple the probabilities of appearance and spatial support matching, and assume that appearance matching is independent of $\mathcal{D}$.
In practice, the appearance term $p(f \mapsto f')$ is simply computed from a similarity between feature descriptors $f$ and $f'$, and the geometric consistency term $p(s \mapsto s' \mid \mathcal{D})$ is evaluated by comparing the spatial supports $s$ and $s'$ in the context of the given data $\mathcal{D}$, as described in the next section. We set the posterior probability as a matching score and assign the best match $\phi(r)$ for each proposal in $\mathcal{R}$:\vspace{-0.1cm}   
\begin{equation}
\label{eq:max}
	\phi(r)=\mathop {\text{argmax} }_{r'\in \mathcal{R^\prime} } p(r \mapsto r' \mid \mathcal{D}).\vspace{-0.1cm}  
\end{equation}
Using a slight abuse of notation, if $(f',s')=\phi(f,s)$, we will write $f'=\phi(f)$ and $s'=\phi(s)$.
\subsection{Geometric matching strategies}
\vspace{-0.1cm}
We now introduce three matching strategies, using different geometric consistency terms $p(s \mapsto s' \mid \mathcal{D})$.
\vspace{-0.4cm}
\paragraph{Naive appearance matching (NAM).}
A straightforward way of matching regions is to use a uniform distribution for the geometric term so that\vspace{-0.1cm}    
\begin{equation}
\label{eq:naive}
	p(r \mapsto r' \mid \mathcal{D}) \propto p(f \mapsto f').\vspace{-0.1cm}   
\end{equation}
NAM considers appearance only, and does not reflect any geometric relationship among regions (Fig. \ref{fig:compare_match}(d)).
\vspace{-0.4cm}
\paragraph{Probabilistic Hough matching (PHM).} 
The matching algorithm in~\cite{cho2015unsupervised} can be expressed in our model as follows. First, a three-dimensional location vector (center position and scale) is extracted from the spatial support $s$. We denote it by a function $\gamma$. An offset space $\mathcal{X}$ is defined as a feasible set of offset vectors between $\gamma(s)$ and $\gamma(s')$: $\mathcal{X}=\{\gamma(s) - \gamma(s') \mid r \in \mathcal{R}, r' \in \mathcal{R}' \}$. The geometric consistency term $p(s \mapsto s' \mid \mathcal{D})$ is then defined as \vspace{-0.1cm} %
\begin{equation}
	p(s \mapsto s' \mid \mathcal{D}) = \sum_{x \in \mathcal{X}} p(s \mapsto s' \mid x) p(x \mid \mathcal{D}),\vspace{-0.1cm} 
\end{equation}
which assumes that $p(s \mapsto s' \mid x, \mathcal{D}) = p(s \mapsto s' \mid x)$. 
Here, $p(s \mapsto s' \mid x)$ measures an offset consistency between $\gamma(s)-\gamma(s')$ and $x$ by a Gaussian kernel in the three-dimensional offset space. From this model, PHM substitutes $p(x \mid \mathcal{D})$ with a generalized Hough transform score:\vspace{-0.1cm}   
\begin{equation}
	h(x \mid \mathcal{D} ) = \sum_{ (r,r') \in \mathcal{D}} p(f \mapsto f') p(s \mapsto s' \mid x).\vspace{-0.1cm}  
\end{equation}
which aggregates individual votes for offset $x$, from \emph{all} possible matches in $\mathcal{D}=\mathcal{R} \times \mathcal{R}'$. Hough voting imposes a spatial regularizer on matching by taking into account a global consensus on the corresponding offset~\cite{leibe2008robust,maji2009object}. However, it often suffers from background clutter that distracts the global voting process (Fig. \ref{fig:compare_match}(e)). 
\vspace{-0.5cm}
\paragraph{Local offset matching (LOM).}
Here we propose a new method to overcome this drawback of PHM~\cite{cho2015unsupervised} and obtain more reliable correspondences. Object proposals often contain a large number of distracting outlier regions from background clutter, and are not perfectly repeatable even for corresponding object or parts across different images (Fig. \ref{fig:compare_match}(c)). The global Hough voting in PHM has difficulties with such outlier regions. In contrast, we optimize a translation and scale offset for each proposal by exploiting only neighboring proposals. That is, instead of averaging $p(s \mapsto s' | x)$ over all feasible offsets $\mathcal{X}$ in PHM, we use one reliable offset optimized for each proposal. This local approach substantially alleviates the effect of outlier regions in matching as will be demonstrated by our experiment results.

The main issue is how to estimate a reliable offset for each proposal $r$ in a robust manner without any information about objects and their locations. One way would be to find the corresponding region of the region $r$ through a multi-scale sliding window search in $\mathcal{I}'$ as in object detection~\cite{felzenszwalb2008discriminatively}, but this is expensive. Instead, we assume that nearby regions have similar offsets. For each region $r$, we first define its neighborhood $\mathcal{N}(r)$ as the regions with overlapping spatial support:\vspace{-0.1cm}  
\begin{equation}
	\mathcal{N}(r)=\{\hat r \mid s \cap \hat s  \neq \emptyset, \hat r \in \mathcal{R} \}. \vspace{-0.1cm}  
\end{equation}
Using an initial correspondence $\phi(r)$, determined by the best match according to the appearance term, each neighboring region $\hat r$ is assigned its own offset, and all of them form a set of neighbor offsets: \vspace{-0.1cm}  
\begin{equation}
	\mathcal{X}(r) =\{ \gamma(\hat s) - \gamma(\phi(\hat s)) \mid \hat r \in \mathcal{N}(r) \}.\vspace{-0.1cm}  
\end{equation} 
 From this set of neighbor offsets, we estimate a local offset $x^*_r$ for the region $r$ by the geometric median~\cite{lopuhaa1991breakdown}\footnote{We found that the centroid and mode of the offset vectors in three-dimensional offset space show worse performance than the geometric median. This is because the neighboring regions may include clutter. Clutter causes incorrect neighbor offsets, but the geometric median is robust to outliers~\cite{fletcher2008robust}, providing a reliable local offset.}:\vspace{-0.1cm}  
\begin{equation}
	x^*_r =\mathop {\text{argmin}}_{x \in \mathbb{R}^3} \sum_{y \in \mathcal{X}(r)} \left\| x-y \right\|_2, \vspace{-0.1cm}  
\end{equation}
which can be globally optimized by Weiszfeld's algorithm~\cite{chandrasekaran1989open} using a form of iteratively re-weighted least squares. 
Based on the local offset $x^*_r$ optimized for each region, we define the geometric consistency function: \vspace{-0.1cm}  
\begin{equation}
	g(s \mapsto s' | \mathcal{D} ) = p(s \mapsto s' | x^*_r)\sum_{\hat r \in \mathcal{N}(r)} p(\hat f \mapsto \phi(\hat f)),\vspace{-0.1cm}  
\end{equation}
which means that $r$ in $\mathcal{R}$ is likely to match with $r'$ in $\mathcal{R}^\prime$ if their offset is close to the local offset $x^*_r$, and $r$ has many neighboring matches with a high appearance fidelity.  

By using $g(s \mapsto s' | \mathcal{D} )$ as a proxy for $p(s \mapsto s' | \mathcal{D} )$, LOM imposes local smoothness on offsets between neighboring regions. This geometric consistency function effectively suppresses matches between clutter regions, while favoring matches between regions that contain objects rather than object parts (Fig.~\ref{fig:compare_match}(f)). In particular, the use of local offsets optimized for each proposal regularizes offsets within a local neighborhood that incorporates an overlap relationship between spatial supports of regions. This local regularization avoids a common problem with PHM, where the matching results often depend on a few strong matches. 

\begin{figure}
\captionsetup{font={small}}
\captionsetup[subfigure]{aboveskip=-0.5pt,belowskip=-0.5pt}
\centering
	\begin{subfigure}{0.475\textwidth}
	\includegraphics[width=\textwidth, frame]{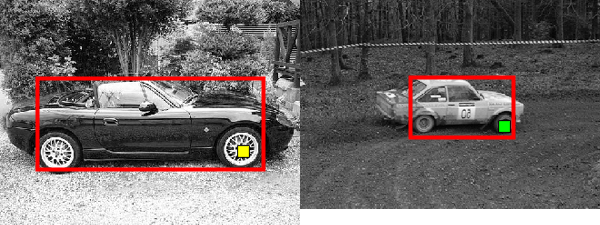}
	\caption{Anchor match and pixel correspondence.}
	\end{subfigure}

	\begin{subfigure}{0.235\textwidth}
	\includegraphics[width=\textwidth,height=0.73\textwidth, frame]{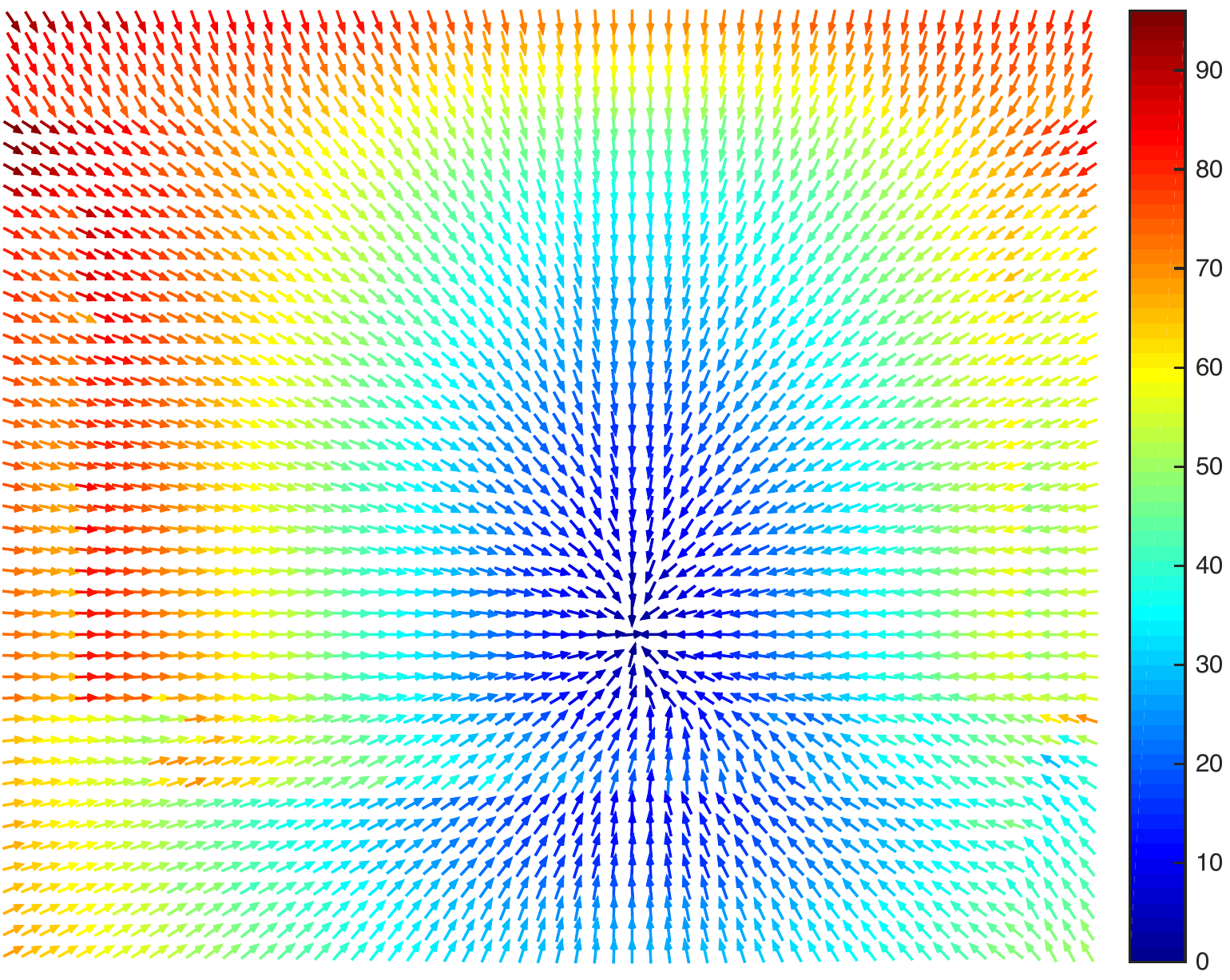}
	\caption{Match visualization.}
	\end{subfigure}
	\begin{subfigure}{0.235\textwidth}
	\includegraphics[width=\textwidth, height=0.73\textwidth, frame]{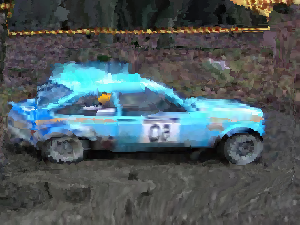}
	\caption{Warped image.}
	\end{subfigure}
	\vfill
	\vspace{-0.2cm}
\caption{Flow field generation. (a) For each pixel (yellow point), its anchor match (red boxes) is determined. The correspondence (green point) is computed by the transformed coordinate with respect to the position and size of the anchor match. (b) Based on the flow field, (c) the right image is warped to the left image. The warped object shows visually similar shape to the one in the left image. The LOM method is used for region matching with the object proposals~\cite{manen2013prime} and the HOG descriptor~\cite{dalal2005histograms}. \bf{(Best viewed in color.)}}
\label{fig:anchor_match}	
\end{figure}
\vspace{-0.2cm}

\subsection{Flow field generation}\label{sec:flowfield}
\vspace{-0.2cm}
The proposal flow gives a set of region correspondences between images, but it can be easily transformed into a conventional dense flow field. 
Let $p$ denote a pixel in image $\mathcal{I}$ (yellow point in Fig.~\ref{fig:anchor_match}(a)). For each pixel $p$, its neighborhood is defined as the region in which it lies, \ie, $\mathcal{N}(p)=\{ r\in \mathcal{R}:p \in r \}$. We define an anchor match $(r^*,\phi(r^*))$ as the region correspondence that has the highest matching score among neighboring regions (red boxes in Fig. \ref{fig:anchor_match}(a)) where \vspace{-0.1cm}  
\begin{equation}\label{eq:anchor_match}
	r^*=\mathop {\text{argmax} }\limits_{r \in \mathcal{N}(p)} p(r \mapsto \phi(r) \mid \mathcal{D}).\vspace{-0.1cm}  
\end{equation}
Note that the anchor match contains information on translation and scale changes between objects. Using the geometric relationships between the pixel $p$ and its anchor match $(r^*,\phi(r^*))$, a correspondence $p'$ in $\mathcal{I}^\prime$ (green point in Fig. \ref{fig:anchor_match}(a)) is obtained by linear interpolation.

The matching score for each correspondence is set to the value of its anchor match. When $p$ and $q$ in $\mathcal{I}$ are matched to the same pixel $p'$ in $\mathcal{I}'$, we select the match with the highest matching score and delete the other one. Finally, joint image filtering~\cite{ham2015robust} is applied under the guidance of the image $\mathcal{I}$ to interpolate the flow field in places without correspondences. Figure \ref{fig:anchor_match}(b-c) shows examples of the estimated flow field and corresponding warping result between two images: Using the dense flow field, we warp all pixels in the right image to the left image. Our approach using the anchor match aligns semantic object parts well while handling translation and scale changes between objects. 
\begin{figure*}
\captionsetup{font={small}}
\captionsetup[subfigure]{aboveskip=-0.5pt,belowskip=-0.5pt}
\centering
	\begin{subfigure}{0.326\textwidth}
	\includegraphics[width=\textwidth, frame]{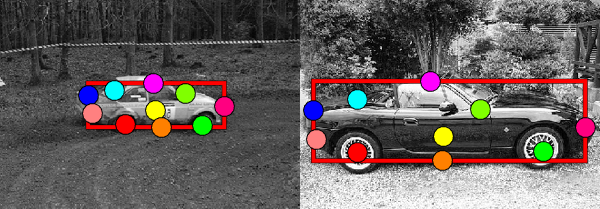}
	\caption{keypoints and object bounding boxs.}
	\end{subfigure}
	\begin{subfigure}{0.163\textwidth}
	\includegraphics[width=\textwidth, frame]{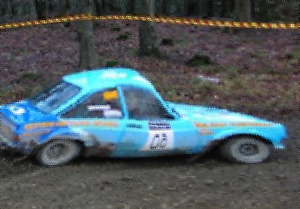}
	\caption{Warping.}
	\end{subfigure}\hspace{-0.07cm}
	\begin{subfigure}{0.163\textwidth}
	\includegraphics[width=\textwidth, frame]{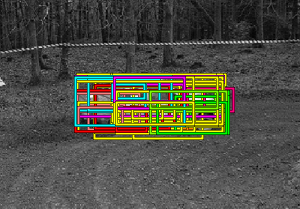}
	\caption{$\mathcal{R}_s$.}
	\end{subfigure}
	\begin{subfigure}{0.326\textwidth}
	\includegraphics[width=\textwidth, frame]{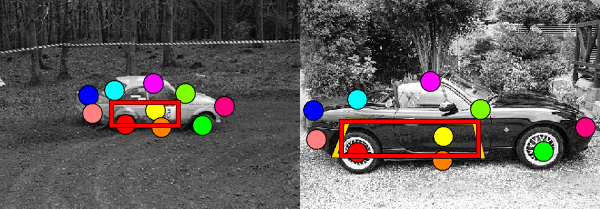}
	\caption{Ground-truth correspondence.}
	\end{subfigure}

	\begin{subfigure}{0.326\textwidth}
	\includegraphics[width=\textwidth, frame]{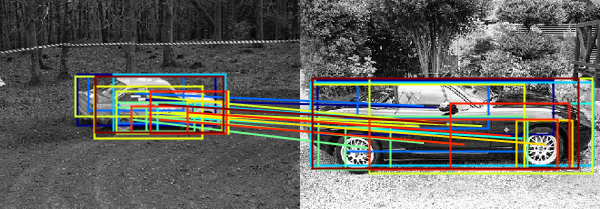}
	\caption{NAM.}
	\end{subfigure}
	\begin{subfigure}{0.326\textwidth}
	\includegraphics[width=\textwidth, frame]{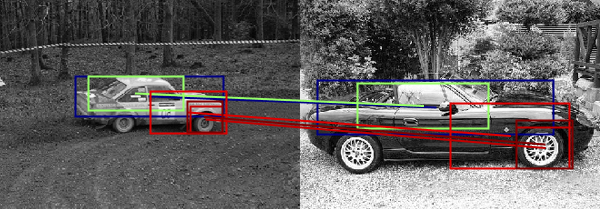}
	\caption{PHM~\cite{cho2015unsupervised}.}
	\end{subfigure}
	\begin{subfigure}{0.326\textwidth}
	\includegraphics[width=\textwidth, frame]{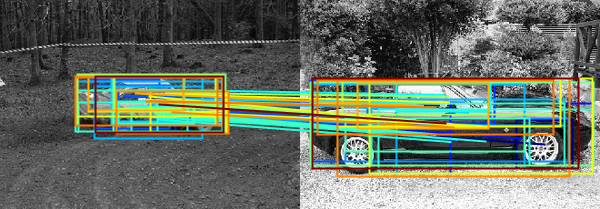}
	\caption{LOM.}
	\end{subfigure}
	\vfill
	\vspace{-0.2cm}
\caption{(a-d) Generating ground-truth regions and evaluating correct
  matches. (a) Using keypoint annotations, dense correspondences
  between images are established using
  warping~\cite{bookstein1989principal, donato2002approximate}. (b)
  Based on the dense correspondences, all pixels in the left image are
  warped to the right image. (c) We assume that true matches exist
  only between the regions near the object bounding box, and thus an
  evaluation is done with the regions in this subset of object
  proposals. (d) For each object proposal (red box in the left image),
  its ground truth is generated automatically by the dense
  correspondences: We use a tight rectangle (red box in the right
  image) of the region formed by the warped object proposal (yellow
  box in the right image) as a ground-truth correspondence. (e-g)
  Examples of correct matches: The numbers of correct matches are 16, 5,
  and 38 for NAM (e), PHM~\cite{cho2015unsupervised} (f), and LOM (g),
  respectively. Matches with IoU score greater than 0.5 are
  considered as correct in this example. \bf{(Best viewed in color.)}}
\label{fig:gt_generation}%
\end{figure*}
\vspace{-0.1cm}
\section{A new dataset for semantic flow evaluation}\label{sec:benchmark} 
\vspace{-0.2cm}
Current research on semantic flow lacks an appropriate benchmark with
dense ground-truth correspondences. Conventional optical flow
benchmarks (e.g., Middlebury~\cite{baker2011database} and
MPI-Sintel~\cite{butler2012naturalistic}) do not feature within-class
variations, and ground truth for generic semantic flow is difficult to
capture due to its intrinsically semantic nature, manual annotation
being extremely labor intensive and somewhat subjective. All existing
approaches are thus evaluated only with sparse ground truth or in an
indirect manner (e.g. mask transfer
accuracy)~\cite{BristowVL15,kim2013deformable,liu2011sift,qiu2014scale,tau2014dense,zhou2015flowweb}.
Such benchmarks only evaluate a small number of matches, that occur at
ground-truth keypoints or around mask boundaries in a point-wise manner. 
To address this issue, we introduce in this section a new dataset for semantic flow, dubbed {\em proposal flow} (PF) dataset, built using ground-truth object bounding boxes and keypoint annotations, (Fig. \ref{fig:gt_generation}(a-b)), and propose new evaluation metrics for region-based semantic flow methods such as proposal flow. Note that while designed for
region-based methods, our benchmark can be used to evaluate any semantic
flow technique. As will be seen in our experiments, it provides a
reasonable (if approximate) ground truth for dense correspondences
across similar scenes without an extremely expensive annotation
campaign. As shown in the following sections, comparative evaluations
on this dataset are also good predictors for performance on other
tasks and datasets, further justifying the use of our benchmark. In the following, we describe our ground-truth
generation process, evaluation criteria, and datasets. The benchmark
data and code are available online:~\url{http://www.di.ens.fr/willow/research/proposalflow}.
\subsection{Ground-truth correspondence generation}
\vspace{-0.1cm}
We assume that true matches only exist within object bounding boxes. Let us assume two sets of keypoint annotations at positions $k_i$ and $k'_i$ in $\mathcal{I}$ and $\mathcal{I}'$, respectively, with $i=1,\ldots,m$. 
Assuming the objects present in the images and their parts may undergo shape deformation, we use thin plate splines (TPS)~\cite{bookstein1989principal, donato2002approximate} to interpolate the sparse keypoints. Namely, the ground truth is approximated from sparse correspondences using TPS warping. 

For each region, its ground-truth match is generated as follows. We assume that true matches only exist between a subset of regions, \ie, regions around object bounding boxes (Fig.~\ref{fig:gt_generation}(c)): $\mathcal{R}_s=\left\{r \mid {\left|b \cap r\right|} \mathbin{/} {\left| r \right|} \geq 0.75, r\in\mathcal{R} \right\}$ where $b$ denotes an object bounding box in $\mathcal{I}$, and $|r|$ indicates the area of a region $r$. %
For each region $r \in \mathcal{R}_s$, the four vertices of the rectangle are warped to the corresponding ones in $\mathcal{I}'$ by the TPS mapping function. The region formed by the warped points is a correspondence of region $r$. 
We fit a tight rectangle for this region and set it as a ground-truth correspondence for the region $r$ (Fig. \ref{fig:gt_generation}(d)). 

\subsection{Evaluation criteria}
\vspace{-0.1cm}
We introduce two evaluation metrics for region matching performance in terms of {\em region matching precision} and {\em match retrieval accuracy}. Basically, the metrics build on the intersection over union (IoU) score between $r$'s correspondence ${\phi(r)}$ and its ground truth $r^\star$:\vspace{-0.1cm} 
\begin{equation} 
	\text{IoU}({\phi(r)},r^\star) = {|{\phi(r)} \cap r^\star|} \mathbin{/} { |{\phi(r)} \cup r^\star|}.\vspace{-0.1cm} 
\end{equation}
For region matching precision, we propose the probability of correct region (PCR) metric\footnote{This region-based metric is based on a conventional point-based metric, the probability of correct keypoint (PCK)~\cite{yang2013articulated}. In the case of pixel-based flow, PCK can be adopted instead.} where region $r$ is correctly matched to its ground truth $r^\star$ if $1-\text{IoU}({\phi(r)},r^\star) < \tau$ (e.g., Fig.~\ref{fig:benchmark}(a) top). We measure the PCR metric while varying the IoU threshold $\tau$ from 0 to 1. For match retrieval accuracy, we propose the average IoU of $k$-best matches (mIoU@$k$) according to the matching score (e.g., Fig. \ref{fig:benchmark}(a) bottom). We measure the mIoU@$k$ metric while increasing the number of top matches $k$. These two metrics exhibit two important characteristics of matching: the PCR reveals the accuracy of overall assignment, and the mIoU@$k$ shows the reliability of matching scores that is crucial in match selection. 
\subsection{Dataset construction}
\vspace{-0.1cm}
To generate our dataset, we start from the benchmark for sparse
matching of Cho {\em et al.}~\cite{cho2013learning}, which consists of 5
object classes (Face, Car, Motorbike, Duck, WineBottle) with 10
keypoint annotations for each image. Note that these images contain
more clutter and intra-class variation than existing datasets
for semantic flow evaluation, \eg, images with tightly cropped objects or
similar
background~\cite{kim2013deformable,qiu2014scale,zhou2015flowweb}. We
exclude the Face class where the number of generated object proposals
is not sufficient to evaluate matching accuracy. The other classes are
split into sub-classes\footnote{They are car (S), (G), (M), duck (S),
  motorbike (S), (G), (M), wine bottle (w/o C), (w/ C), (M), where (S)
  and (G) denote side and general viewpoints, respectively. (C) stands
  for background clutter, and (M) denotes mixed viewpoints (side +
  general) for car and motorbike classes and a combination of images
  in wine bottle (w/o C + w/ C) for the wine bottle class. The dataset
  has 10 images for each class, thus 100 images in total.} according
to viewpoint or background clutter. We obtain a total of 10
sub-classes.  Given these images and regions, we generate ground-truth
data between all possible image pairs within each class.  %
\section{Experiments}
\vspace{-0.1cm}
\subsection{Experimental details}
\vspace{-0.1cm}
\paragraph{Object proposals.}
We evaluate four state-of-the-art object proposal methods:~EdgeBox~(EB)~\cite{zitnick2014edge},~multi-scale combinatorial grouping~(MCG)~\cite{arbelaez2014multiscale},~selective search~(SS)~\cite{uijlings2013selective},~and randomized prim~(RP)~\cite{manen2013prime}.~In addition, we consider three baseline proposals~\cite{hosang2015what}:~Uniform sampling~(US), Gaussian sampling~(GS), and sliding window~(SW). See~\cite{hosang2015what} for more details. For fair comparison, we use 1,000 proposals for all the methods. To control the number of proposals, we use the proposal score provided by EB, MCG, and SS. For RP, we randomly select among the proposals.
\vspace{-0.5cm}
\paragraph{Feature descriptors and similarity.}
We evaluate three popular feature descriptors: SPM~\cite{lazebnik2006beyond}, HOG~\cite{dalal2005histograms}, and ConvNet~\cite{krizhevsky2012imagenet}. For SPM, dense SIFT features~\cite{lowe2004distinctive} are extracted every 4 pixels and each descriptor is quantized into a 1,000 word codebook~\cite{tang2014co}. For each region, a spatial pyramid pooling~\cite{lazebnik2006beyond} is used with $1\times1$ and $3\times3$ pooling regions. We compute the similarity between SPM descriptors by the $\chi^2$ kernel. HOG features are extracted with $8\times8$ cells and 31 orientations, then whitened. For ConvNet features, we use each output of the 5 convolutional layers in AlexNet~\cite{krizhevsky2012imagenet}, which is pre-trained on the ImageNet dataset~\cite{deng2009imagenet}. For HOG and ConvNet, the dot product is used as a similarity metric.
\subsection{Proposal flow components}\label{sec:ex-region-matching}
\vspace{-0.1cm}
We use the PF benchmark in this section to compare three variants of
proposal flow using different
matching algorithms (NAM, PHM, LOM), combined with various object
proposals~\cite{arbelaez2014multiscale,hosang2015what,manen2013prime,uijlings2013selective,zitnick2014edge},
and
features~\cite{dalal2005histograms,krizhevsky2012imagenet,lazebnik2006beyond}.
\begin{figure*}
\captionsetup{font={small}}
\captionsetup[subfigure]{aboveskip=-0.1pt,belowskip=-0.1pt}
\centering
	\begin{subfigure}{0.32\textwidth}
	\includegraphics[width=\textwidth]{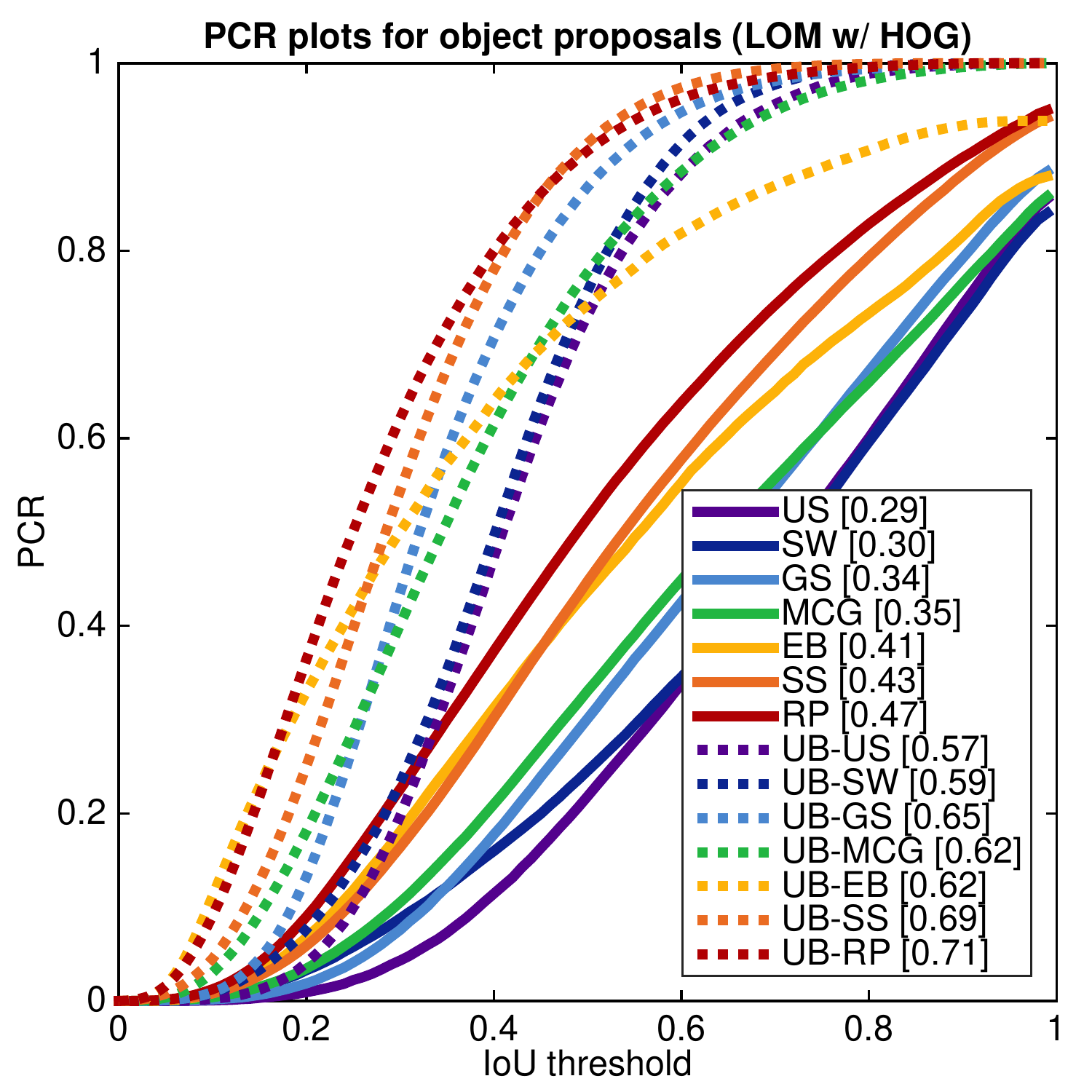}
	\end{subfigure}
	\begin{subfigure}{0.32\textwidth}
	\includegraphics[width=\textwidth]{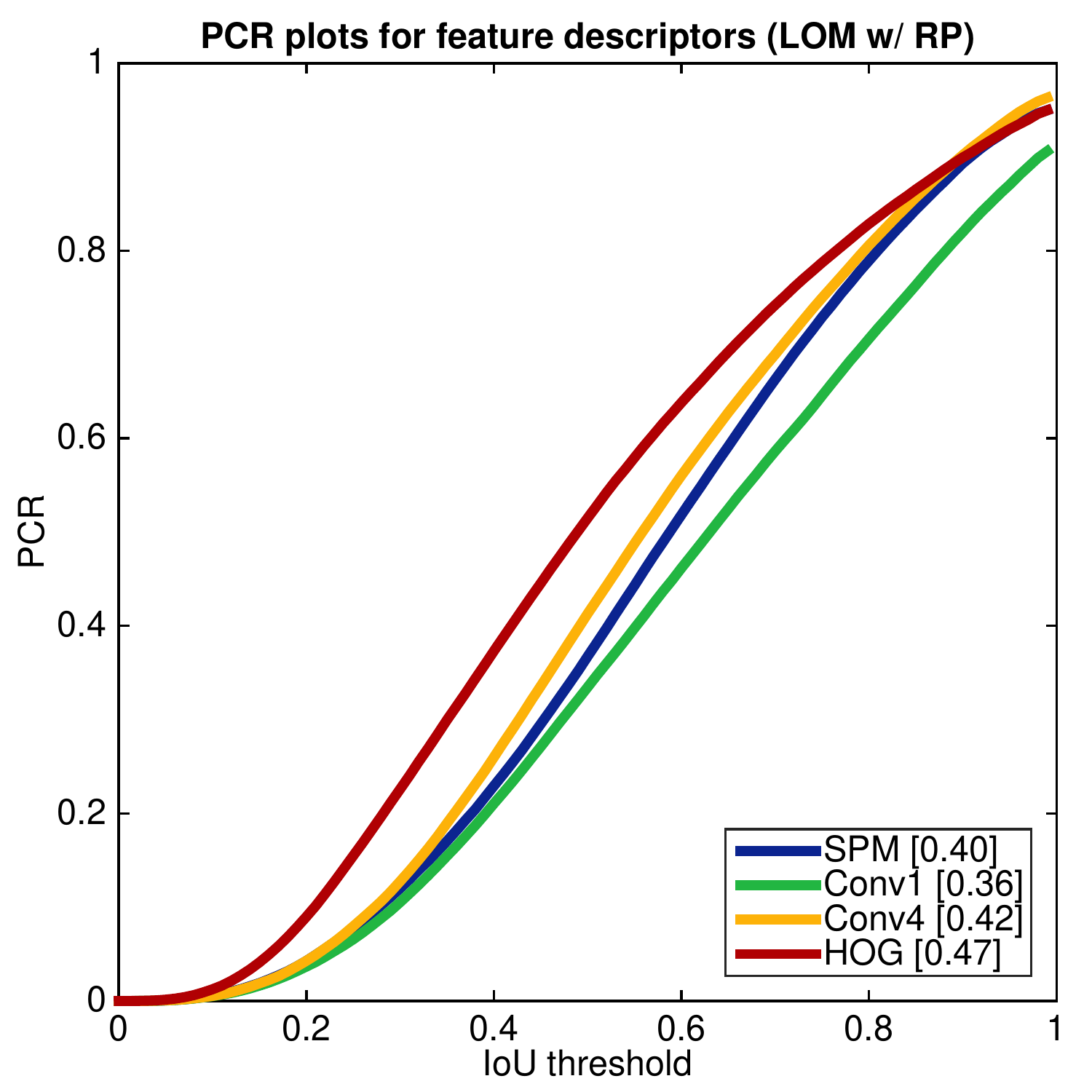}
	\end{subfigure}
	\begin{subfigure}{0.32\textwidth}
	\includegraphics[width=\textwidth]{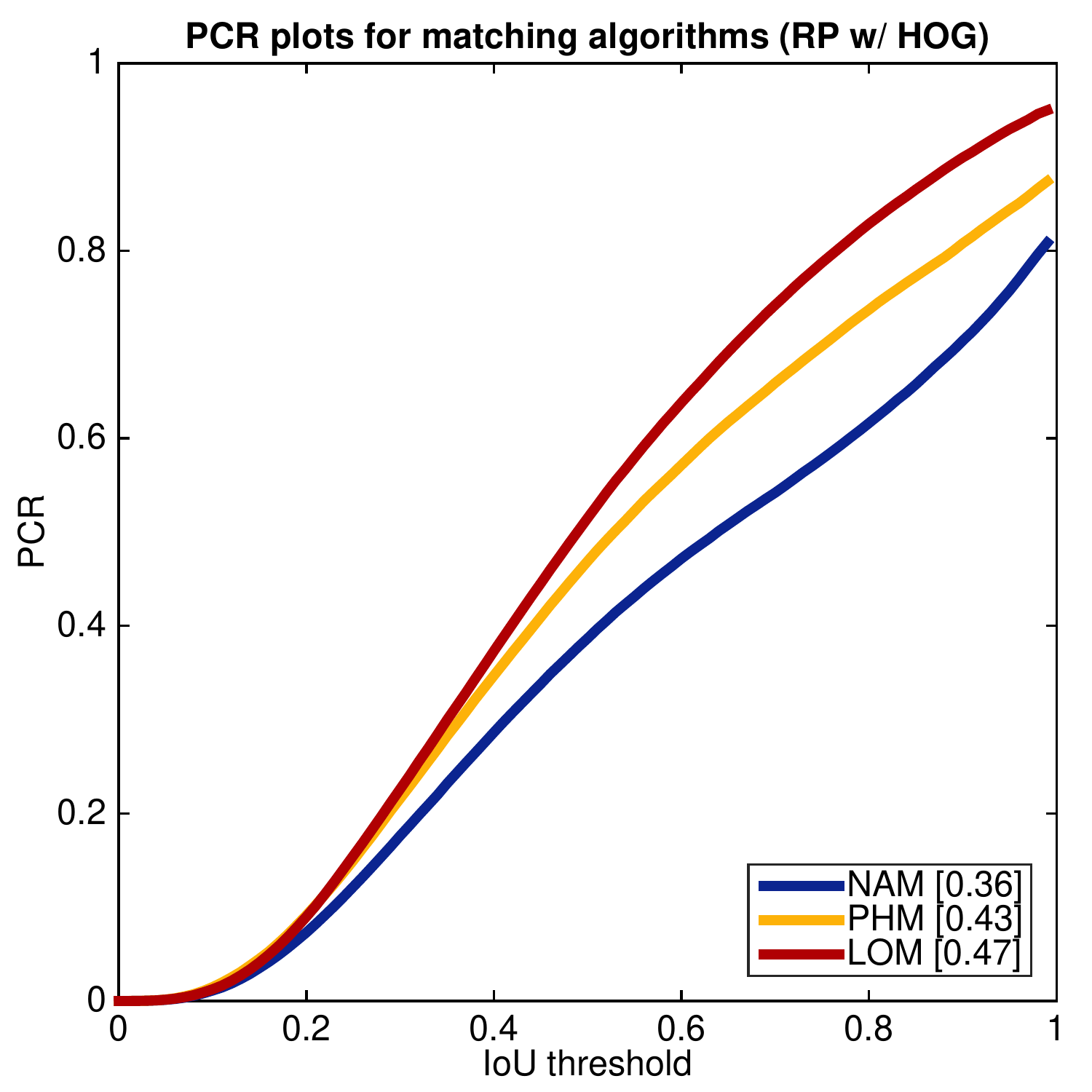}
	\end{subfigure}

	\begin{subfigure}{0.32\textwidth}
	\includegraphics[width=\textwidth]{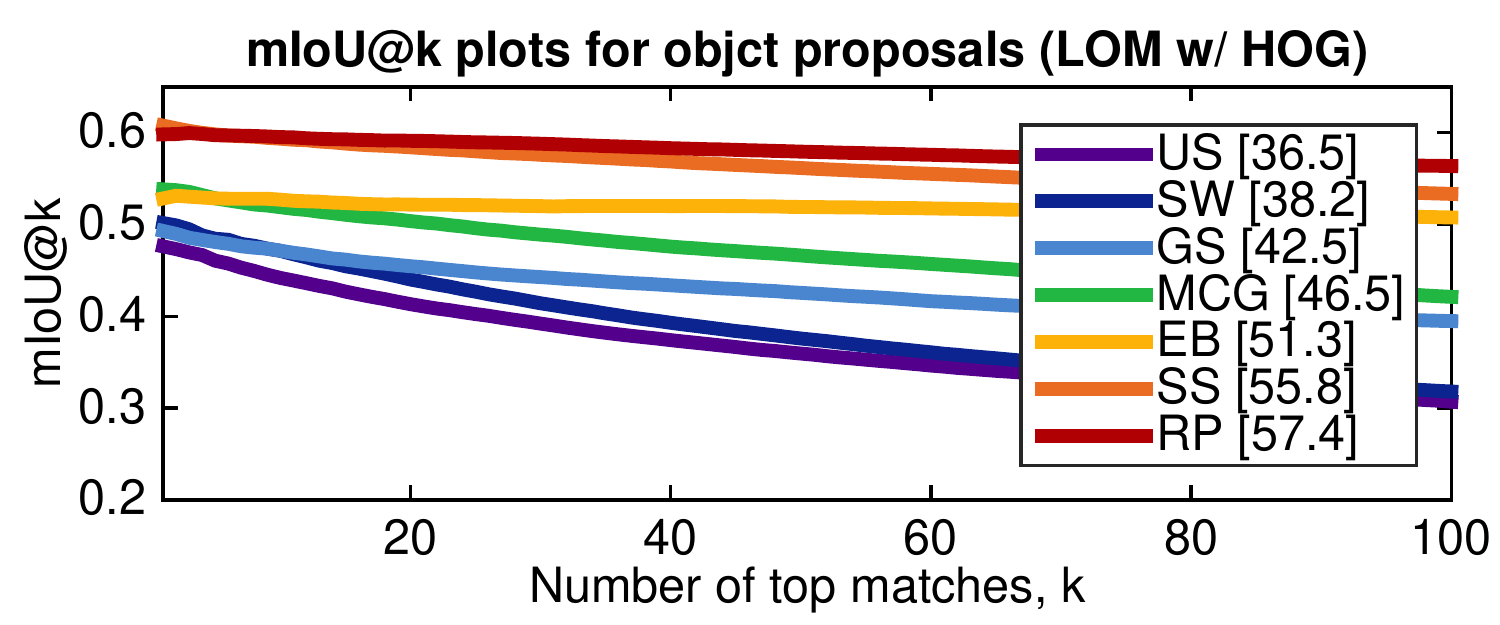}
	\caption{Comparison of object proposals.}
	\end{subfigure}
	\begin{subfigure}{0.32\textwidth}
	\includegraphics[width=\textwidth]{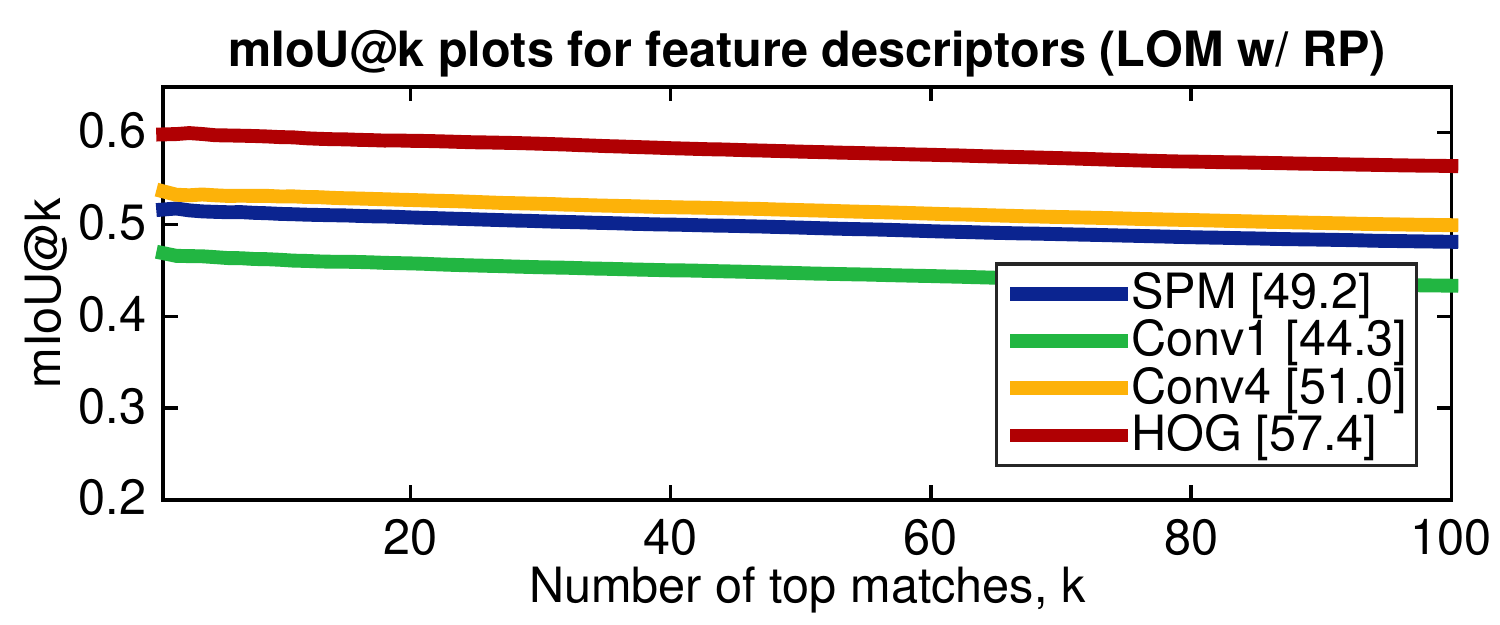}
	\caption{Comparison of feature descriptors.}
	\end{subfigure}
	\begin{subfigure}{0.32\textwidth}
	\includegraphics[width=\textwidth]{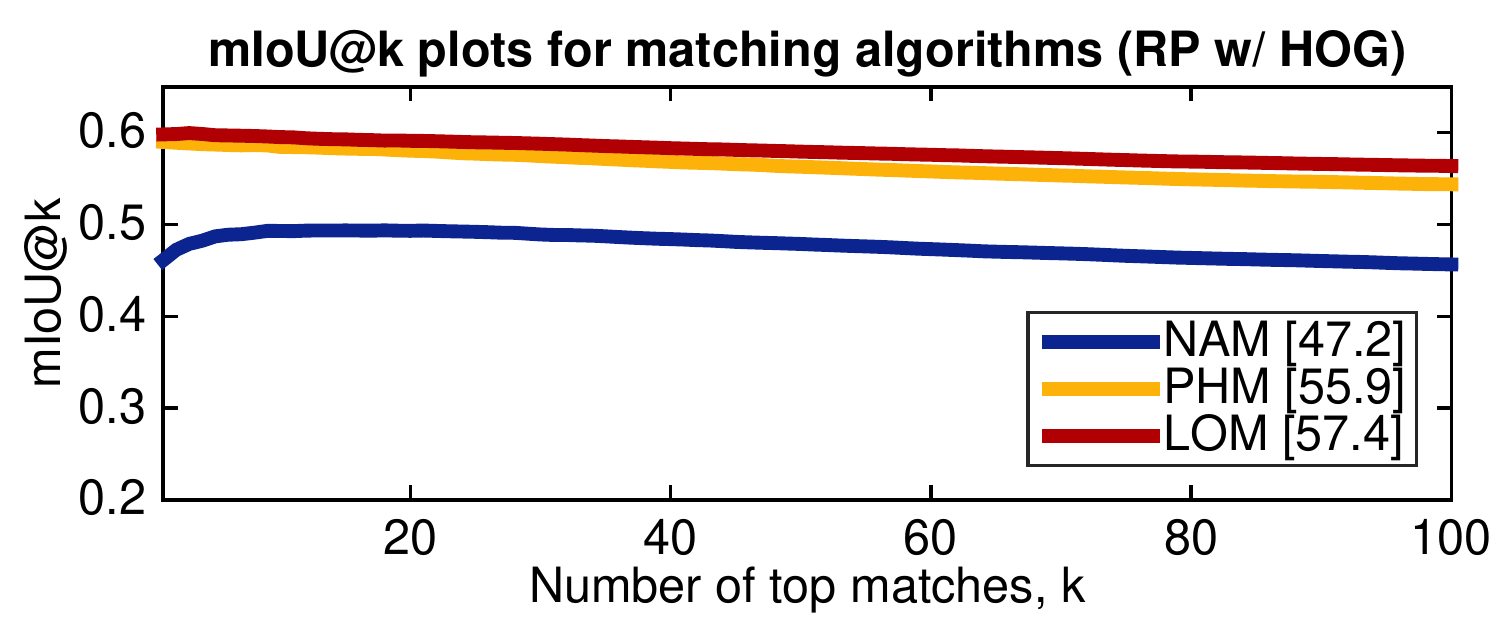}
	\caption{Comparison of matching algorithms.}
	\end{subfigure}

    \begin{subfigure}{0.95\textwidth}
	\includegraphics[width=\textwidth]{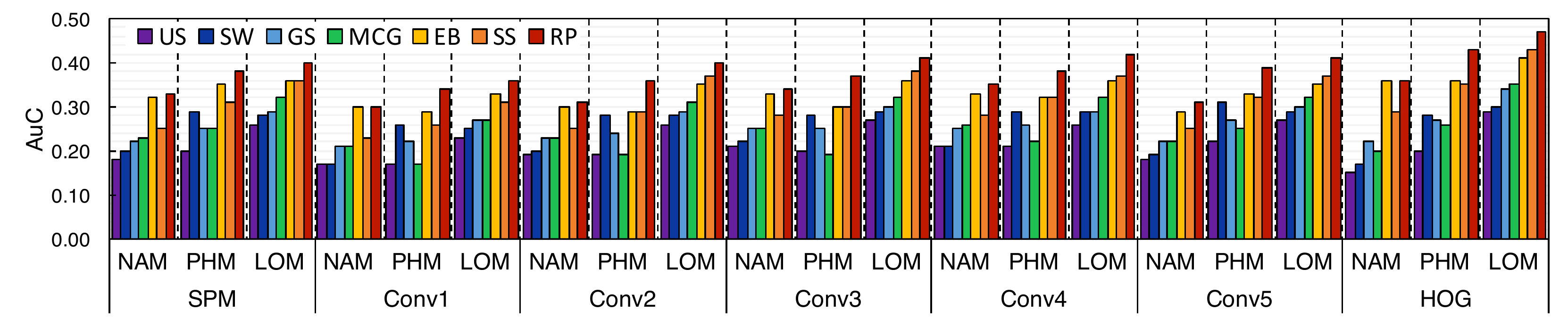}
	\caption{AuCs for PCR curves.}
	\end{subfigure}

	\begin{subfigure}{0.95\textwidth}
	\includegraphics[width=\textwidth]{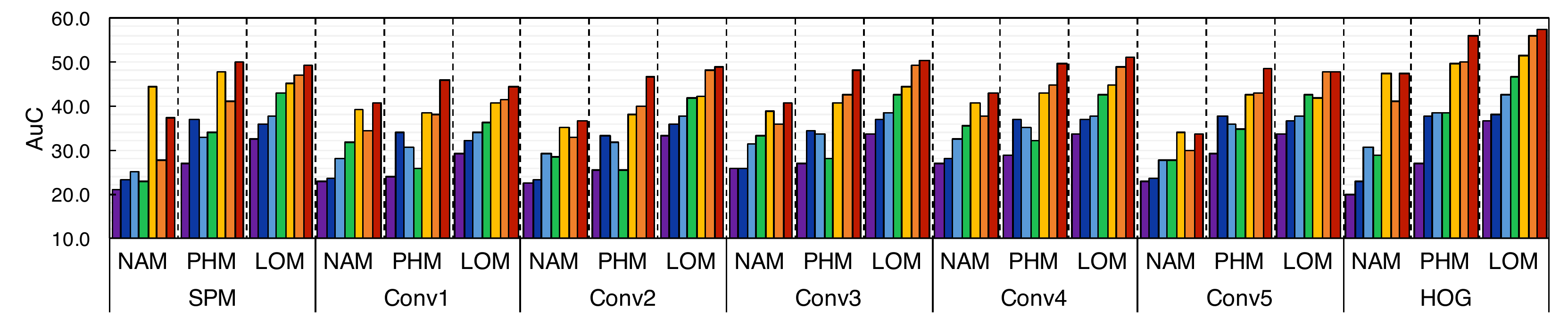}
	\caption{AuCs for mIoU@$k$ curves.}
	\end{subfigure}
	\vfill
	\vspace{-0.2cm}

\caption{PF benchmark evaluation on (a-c) region matching precision (top, PCR plots) and match retrieval accuracy (bottom, mIoU@$k$ plots), and (d-e) AuCs for different combinations of object proposals, feature descriptors, and matching algorithms: (a) Evaluation for LOM with HOG~\cite{dalal2005histograms}, (b) evaluation for LOM with RP~\cite{manen2013prime}, (c) evaluation for RP with HOG~\cite{dalal2005histograms}, (d) AuCs for PCR plots, and (e) AuCs for mIoU@$k$ plots. The AuC is shown in the legend. \bf{(Best viewed in color.)}}
\label{fig:benchmark}	
\end{figure*}

Figure \ref{fig:gt_generation}(e-g) shows a qualitative comparison between region matching algorithms on a pair of images and depicts correct matches found by each variant of proposal flow. In this example, at the IoU threshold $0.5$, the numbers of correct matches are 16, 5, and 38 for NAM, PHM~\cite{cho2015unsupervised}, and LOM, respectively. This shows that PHM may give worse performance than even NAM when much clutter exists in background. In contrast, the local regularization in LOM alleviates the effect of such clutter.

Figure \ref{fig:benchmark} summarizes the matching and retrieval performance on average for all object classes with a variety of combination of object proposals, feature descriptors, and matching algorithms. Figure \ref{fig:benchmark}(a) compares different types of object proposals with fixed matching algorithm and feature descriptor (LOM w/ HOG). RP shows the best matching precision and retrieval accuracy among the object proposals. %
An upper bound on precision is measured for object proposals (around a given object) in the image $\mathcal{I}$ using a corresponding ground truths in image $\mathcal{I}^\prime$, that is the best matching accuracy we can achieve with each proposal method. The upper bound (UB) plots show that RP generates more consistent regions than other proposal methods, and is adequate for region matching. RP shows higher matching precision than other proposals especially when the IoU threshold is low. %
The evaluation results for different features (LOM w/ RP) are shown in Fig.~\ref{fig:benchmark}(b). The HOG descriptor gives the best performance in matching and retrieval. The CNN features in our comparison come from AlexNet~\cite{krizhevsky2012imagenet} trained for ImageNet classification.~Such CNN features have a task-specific bias to capture discriminative parts for classification, which may be less adequate for patch correspondence or retrieval than engineered features such as HOG. Similar conclusions are found in recent papers~\cite{long2014convnets, paulin2015localetal}. See, for example, Table 3 in~\cite{paulin2015localetal} where SIFT outperforms all AlexNet features (Conv1-5). Among ConvNet features, the fourth and first convolutional layers (Conv4 and Conv1) show the best and worst performance, respectively, while other layers perform similar to SPM. This confirms the finding in~\cite{zagoruyko2015learning}, which shows that Conv4 gives the best matching performance among ImageNet-trained ConvNet features. Figure \ref{fig:benchmark}(c) compares the performance of different matching algorithms (RP w/ HOG), and shows that LOM outperforms others in matching as well as retrieval. Figure \ref{fig:benchmark}(d and e) shows the area under curve (AuC) for PCR and mIoU@$k$ plots, respectively. This suggests that combining LOM, RP, and HOG performs best in both metrics.

 In Table \ref{tb:iou_per_cat}, we show AuCs of PCR plots for each class (LOM w/ RP and HOG). From this table, we can see that 1) higher matching precision is achieved with objects having a similar pose (e.g., mot(S)~vs.~mot(M)), 2) performance decreases for deformable object matching (e.g., duck(S)~vs.~car(S)), and 3) matching precision can increase drastically by eliminating background clutters (e.g., win(w/o C)~vs.~win(w/ C)). 
\begin{table*}
\centering
\footnotesize
\captionsetup{font={small}}
\caption{AuC performance for PCR plots on the PF dataset (LOM w/ RP and HOG).}
\vspace{-0.3cm}
\addtolength{\tabcolsep}{-2.0pt}
\begin{tabular}{l c c c c c c c c c c c}
\toprule
\multicolumn{1}{c}{Methods} & car(S) & car(G) & car(M) & duck(S) & mot(S) & mot(G) & mot(M) & win(w/o C) & win(w/ C) & win(M) & Avg. \\
\midrule
\midrule
LOM	                         & 0.61 & 	0.50	&   0.45&	0.50	&  0.42& 	0.40&	0.35&	0.69&	0.30&	0.47&	0.47\\	
Upper bound & 0.75&	0.69&	0.69&	0.72&	0.70&	0.70&	0.67&	0.80&	0.68&	0.73&	0.71\\
\bottomrule
\end{tabular}%
\label{tb:iou_per_cat}
\end{table*}
\begin{table}
\centering
\footnotesize
\captionsetup{font={small}}
\caption{PCK comparison for dense flow field on the PF dataset.}
\vspace{-0.3cm}
\addtolength{\tabcolsep}{-2.0pt}
\begin{tabular}{l c c c c}
\toprule
\multicolumn{1}{c}{Methods} & MCG~\cite{arbelaez2014multiscale} & EB~\cite{zitnick2014edge} & SS~\cite{uijlings2013selective} & RP~\cite{manen2013prime}\\
\midrule
\midrule
NAM & 0.46 & \textbf{0.50}&  0.52 & 0.53\\
PHM  & 0.48&	0.45&0.55&0.54\\
LOM &	 \textbf{0.49}&0.44&\textbf{0.56}&\textbf{0.55}\\
\midrule
DeepFlow~\cite{weinzaepfel2015deepmatching} & \multicolumn{4}{c}{0.20}\\
GMK~\cite{duchenne2011graph} & \multicolumn{4}{c}{0.27}\\
SIFT Flow~\cite{liu2011sift} & \multicolumn{4}{c}{0.38}\\
DSP~\cite{kim2013deformable} & \multicolumn{4}{c}{0.37}\\
\bottomrule
\end{tabular}
\label{tb:pck}
\end{table}
\begin{figure*}
\captionsetup{font={small}}
\captionsetup[subfigure]{aboveskip=-0.5pt,belowskip=-0.5pt}
\centering
	\begin{subfigure}{0.135\textwidth}
	\includegraphics[width=\textwidth, frame]{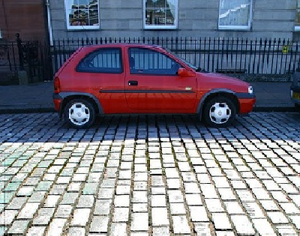}
	\end{subfigure}
	\begin{subfigure}{0.135\textwidth}
	\includegraphics[width=\textwidth, frame]{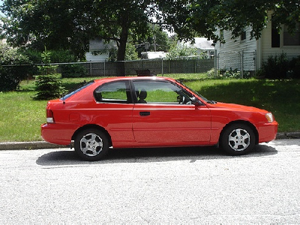}
	\end{subfigure}
	\begin{subfigure}{0.135\textwidth}
	\includegraphics[width=\textwidth, frame]{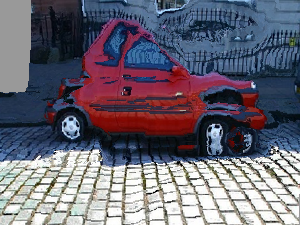}
	\end{subfigure}
	\begin{subfigure}{0.135\textwidth}
	\includegraphics[width=\textwidth, frame]{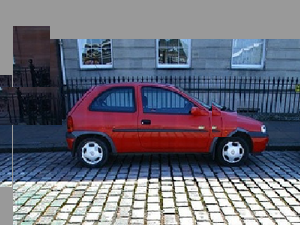}
	\end{subfigure}
	\begin{subfigure}{0.135\textwidth}
	\includegraphics[width=\textwidth, frame]{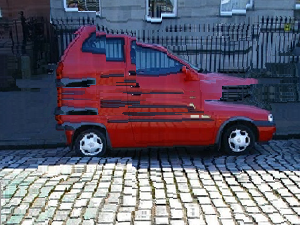}
	\end{subfigure}
	\begin{subfigure}{0.135\textwidth}
	\includegraphics[width=\textwidth, frame]{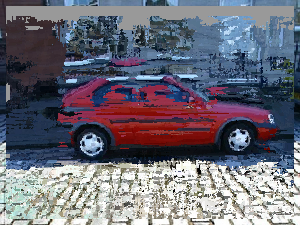}
	\end{subfigure}
	\begin{subfigure}{0.135\textwidth}
	\includegraphics[width=\textwidth, frame]{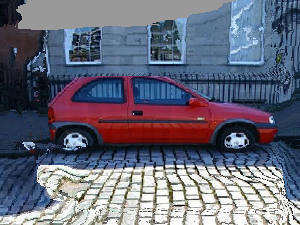}
	\end{subfigure}
	
	\begin{subfigure}{0.135\textwidth}
	\includegraphics[width=\textwidth, frame]{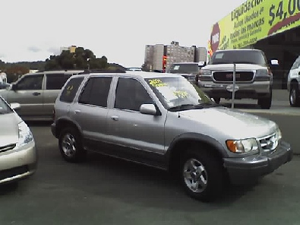}
	\end{subfigure}
	\begin{subfigure}{0.135\textwidth}
	\includegraphics[width=\textwidth, frame]{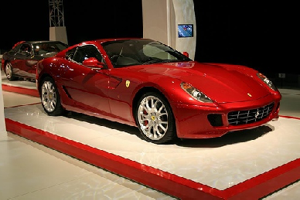}
	\end{subfigure}
	\begin{subfigure}{0.135\textwidth}
	\includegraphics[width=\textwidth, frame]{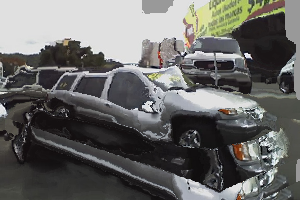}
	\end{subfigure}
	\begin{subfigure}{0.135\textwidth}
	\includegraphics[width=\textwidth, frame]{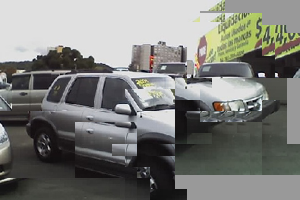}
	\end{subfigure}
	\begin{subfigure}{0.135\textwidth}
	\includegraphics[width=\textwidth, frame]{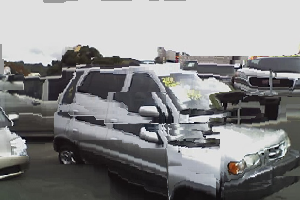}
	\end{subfigure}
	\begin{subfigure}{0.135\textwidth}
	\includegraphics[width=\textwidth, frame]{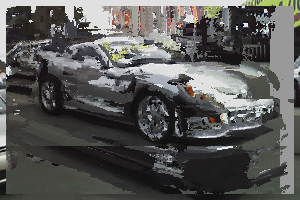}
	\end{subfigure}
	\begin{subfigure}{0.135\textwidth}
	\includegraphics[width=\textwidth, frame]{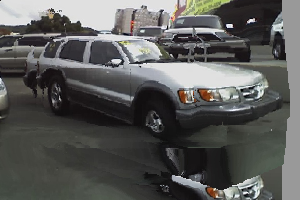}
	\end{subfigure}
	
	\begin{subfigure}{0.135\textwidth}
	\includegraphics[width=\textwidth, frame]{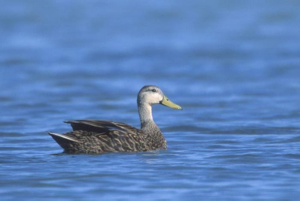}
	\end{subfigure}
	\begin{subfigure}{0.135\textwidth}
	\includegraphics[width=\textwidth, frame]{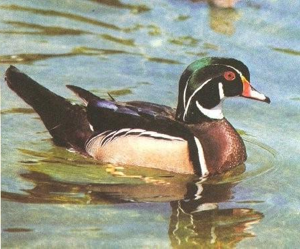}
	\end{subfigure}
	\begin{subfigure}{0.135\textwidth}
	\includegraphics[width=\textwidth, frame]{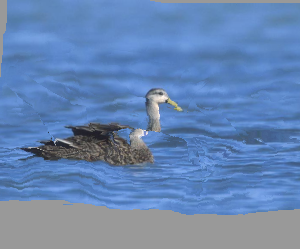}
	\end{subfigure}
	\begin{subfigure}{0.135\textwidth}
	\includegraphics[width=\textwidth, frame]{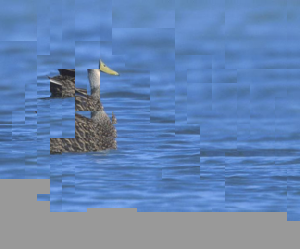}
	\end{subfigure}
	\begin{subfigure}{0.135\textwidth}
	\includegraphics[width=\textwidth, frame]{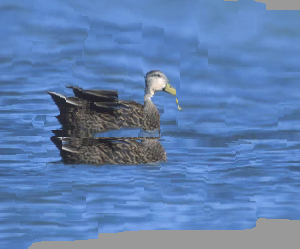}
	\end{subfigure}
	\begin{subfigure}{0.135\textwidth}
	\includegraphics[width=\textwidth, frame]{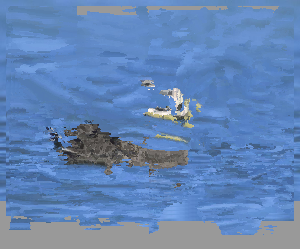}
	\end{subfigure}
	\begin{subfigure}{0.135\textwidth}
	\includegraphics[width=\textwidth, frame]{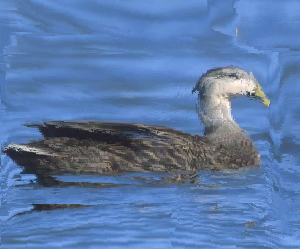}
	\end{subfigure}
	
	\begin{subfigure}{0.135\textwidth}
	\includegraphics[width=\textwidth, frame]{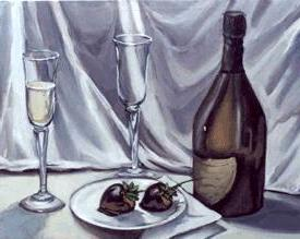}
	\end{subfigure}
	\begin{subfigure}{0.135\textwidth}
	\includegraphics[width=\textwidth, frame]{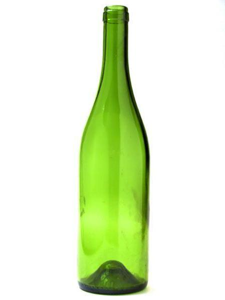}
	\end{subfigure}
	\begin{subfigure}{0.135\textwidth}
	\includegraphics[width=\textwidth, frame]{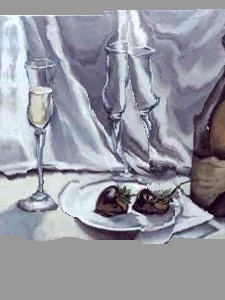}
	\end{subfigure}
	\begin{subfigure}{0.135\textwidth}
	\includegraphics[width=\textwidth, frame]{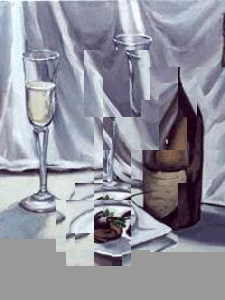}
	\end{subfigure}
	\begin{subfigure}{0.135\textwidth}
	\includegraphics[width=\textwidth, frame]{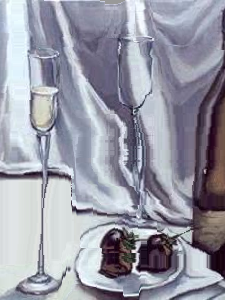}
	\end{subfigure}
	\begin{subfigure}{0.135\textwidth}
	\includegraphics[width=\textwidth, frame]{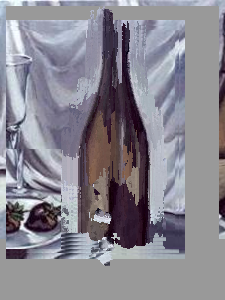}
	\end{subfigure}
	\begin{subfigure}{0.135\textwidth}
	\includegraphics[width=\textwidth, frame]{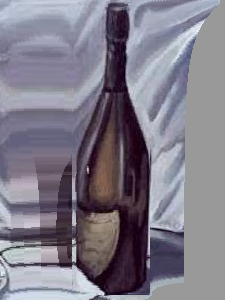}
	\end{subfigure}

	\begin{subfigure}{0.135\textwidth}
	\includegraphics[width=\textwidth, frame]{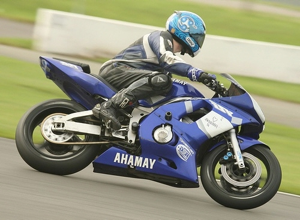}
	\end{subfigure}
	\begin{subfigure}{0.135\textwidth}
	\includegraphics[width=\textwidth, frame]{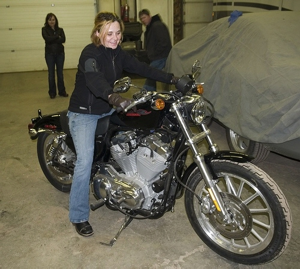}
	\end{subfigure}
	\begin{subfigure}{0.135\textwidth}
	\includegraphics[width=\textwidth, frame]{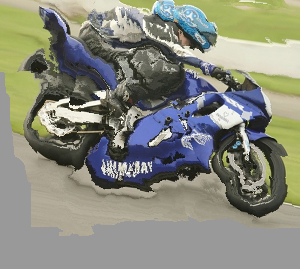}
	\end{subfigure}
	\begin{subfigure}{0.135\textwidth}
	\includegraphics[width=\textwidth, frame]{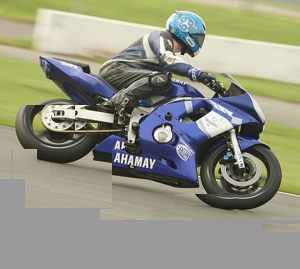}
	\end{subfigure}
	\begin{subfigure}{0.135\textwidth}
	\includegraphics[width=\textwidth, frame]{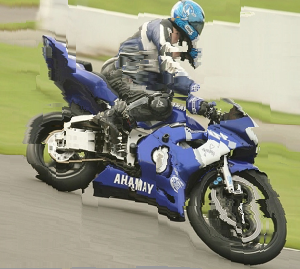}
	\end{subfigure}
	\begin{subfigure}{0.135\textwidth}
	\includegraphics[width=\textwidth, frame]{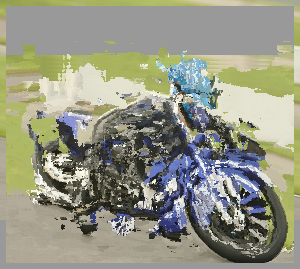}
	\end{subfigure}
	\begin{subfigure}{0.135\textwidth}
	\includegraphics[width=\textwidth, frame]{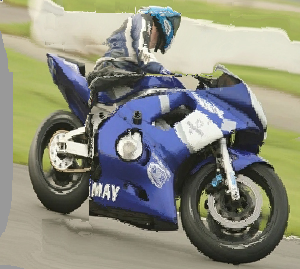}
	\end{subfigure}

	\begin{subfigure}{0.135\textwidth}
	\includegraphics[width=\textwidth, frame]{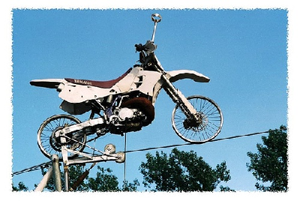}
	\caption{Source image.}
	\end{subfigure}
	\begin{subfigure}{0.135\textwidth}
	\includegraphics[width=\textwidth, frame]{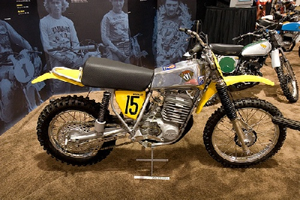}
	\caption{Target image.}
	\end{subfigure}
	\begin{subfigure}{0.135\textwidth}
	\includegraphics[width=\textwidth, frame]{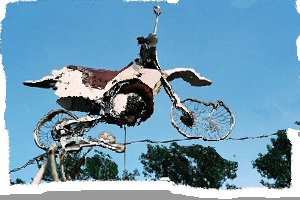}
	\caption{DeepFlow.}
	\end{subfigure}
	\begin{subfigure}{0.135\textwidth}
	\includegraphics[width=\textwidth, frame]{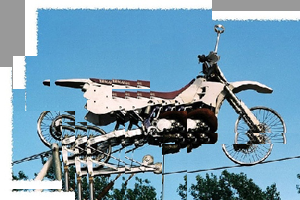}
	\caption{GMK.}
	\end{subfigure}
	\begin{subfigure}{0.135\textwidth}
	\includegraphics[width=\textwidth, frame]{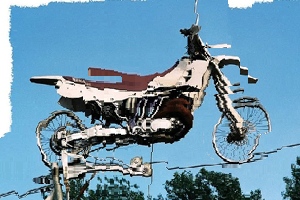}
	\caption{SIFT Flow.}
	\end{subfigure}
	\begin{subfigure}{0.135\textwidth}
	\includegraphics[width=\textwidth, frame]{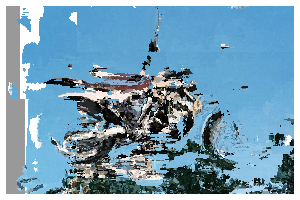}
	\caption{DSP.}
	\end{subfigure}
	\begin{subfigure}{0.135\textwidth}
	\includegraphics[width=\textwidth, frame]{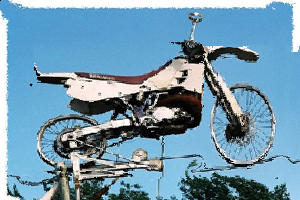}
	\caption{Proposal Flow.}
	\end{subfigure}
	\vfill
	\vspace{-0.2cm}
\caption{Examples of dense flow field.~(a-b) Sourse images are warped to the target images using the dense correspondences estimated by (c) DeepFlow~\cite{weinzaepfel2015deepmatching}, (d) GMK~\cite{duchenne2011graph}, (e) SIFT Flow~\cite{liu2011sift}, (f) DSP~\cite{kim2013deformable}, and (g) Proposal Flow (LOM w/ RP and HOG).}
\label{fig:correspondence}	
\end{figure*}
\vspace{-0.2cm}
\subsection{Flow field}
\vspace{-0.1cm}
To compare our method with state-of-the-art semantic flow methods, we compute a dense flow field from our proposal flows (Sec.~\ref{sec:flowfield}), and evaluate image alignment between all pairs of images in each subset of the PF dataset. We test four object proposal methods (MCG, EB, SS, RP) with HOG descriptors. For an evaluation metric, we use PCK between warped keypoints and ground-truth ones~\cite{long2014convnets, yang2013articulated}. Ground-truth keypoints are deemed to be correctly predicted if they lie within $\alpha \text{max}(h,w)$ pixels of the predicted points for $\alpha$ in $[0, 1]$, where $h$ and $w$ are the height and width of the object bounding box, respectively.
Table \ref{tb:pck} shows the average PCK $(\alpha = 0.1)$ over all object classes. In our benchmark, all versions of proposal flow significantly outperform SIFT Flow~\cite{liu2011sift}, DSP~\cite{kim2013deformable}, and DeepFlow~\cite{weinzaepfel2015deepmatching}. LOM with SS or RP outperforms other combination of matching and proposal methods, which coincides with the results in Sec \ref{sec:ex-region-matching}. Figure \ref{fig:correspondence} gives a qualitative comparison with the state of the art on the PF dataset. The better alignment found by proposal flow here is typical of our experiments. Specifically, proposal flow is robust to translation and scale changes between objects. 
\begin{table}
\centering
\footnotesize
\captionsetup{font={small}}
\caption{Matching accuracy on the Caltech-101 dataset.}
\vspace{-0.3cm}
\addtolength{\tabcolsep}{-2.0pt}
\begin{tabular}{ c l c c c }
\toprule
Proposals & \multicolumn{1}{c}{Methods} & LT-ACC & IoU & LOC-ERR \\
\midrule
\midrule
\multirow{3}{*}{SS~\cite{uijlings2013selective}} & NAM & 0.68 & 0.44 & 0.41 \\
                    & PHM & 0.74 & 0.48 & 0.32 \\
					& LOM & \textbf{0.78} & \textbf{0.50} & \textbf{0.25} \\
\midrule
\multirow{3}{*}{RP~\cite{manen2013prime}} & NAM & 0.70 & 0.44 & 0.39\\
					& PHM & 0.75 & 0.48 & 0.31\\
					& LOM & \textbf{0.78} & \textbf{0.50} & 0.26\\
\midrule
\multicolumn{2}{l}{DeepFlow~\cite{weinzaepfel2015deepmatching}}  &0.74& 0.40& 0.34\\
\multicolumn{2}{l}{GMK~\cite{duchenne2011graph}}  & 0.77& 0.42&0.34\\
\multicolumn{2}{l}{SIFT Flow~\cite{liu2011sift}}        & 0.75 & 0.48 & 0.32\\
\multicolumn{2}{l}{DSP~\cite{kim2013deformable}}              & 0.77 & 0.47 & 0.35\\
\bottomrule
\end{tabular}
\label{tb:caltech}
\end{table}
\begin{table}
\centering
\footnotesize
\captionsetup{font={small}}
\caption{Matching accuracy on the PASCAL VOC classes.}
\vspace{-0.3cm}
\addtolength{\tabcolsep}{-2.0pt}
\begin{tabular}{c l c c }
\toprule
Proposals & \multicolumn{1}{c}{Methods} & IoU & PCK\\
\midrule
\midrule
\multirow{3}{*}{SS~\cite{uijlings2013selective}} & NAM								& 0.35	&0.13\\
& PHM								& 0.39	&\textbf{0.17}\\
& LOM 								& \textbf{0.41} &\textbf{0.17}\\
\midrule
\multicolumn{2}{l}{Congealing~\cite{learned2006data}} 				& 0.38 	&0.11\\		    
\multicolumn{2}{l}{RASL~\cite{peng2012rasl}}          				& 0.39	&0.16\\
\multicolumn{2}{l}{CollectionFlow~\cite{kemelmacher2012collection}} & 0.38	&0.12\\
\multicolumn{2}{l}{DSP~\cite{kim2013deformable}}						& 0.39	&\textbf{0.17}\\
\midrule
\multicolumn{2}{l}{FlowWeb~\cite{zhou2015flowweb}}						& \textbf{0.43}	&\textbf{0.26}\\
\bottomrule
\end{tabular}
\label{tb:voc}
\end{table}

\begin{figure*}[ht!]
\captionsetup{font={small}}
\captionsetup[subfigure]{aboveskip=-0.5pt,belowskip=-0.5pt}
\centering
	\begin{subfigure}{0.12\textwidth}
	\includegraphics[width=\textwidth]{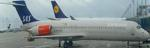}
	\end{subfigure}
	\begin{subfigure}{0.12\textwidth}
	\includegraphics[width=\textwidth]{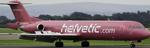}
	\end{subfigure}
	\begin{subfigure}{0.12\textwidth}
	\includegraphics[width=\textwidth]{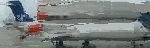}
	\end{subfigure}
	\begin{subfigure}{0.12\textwidth}
	\includegraphics[width=\textwidth]{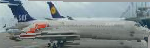}
	\end{subfigure}
	\begin{subfigure}{0.12\textwidth}
	\includegraphics[width=\textwidth]{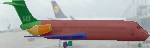}
	\end{subfigure}
	\begin{subfigure}{0.12\textwidth}
	\includegraphics[width=\textwidth]{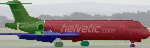}
	\end{subfigure}
	\begin{subfigure}{0.12\textwidth}
	\includegraphics[width=\textwidth]{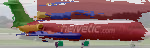}
	\end{subfigure}
	\begin{subfigure}{0.12\textwidth}
	\includegraphics[width=\textwidth]{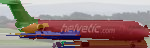}
	\end{subfigure}
		
	\begin{subfigure}{0.12\textwidth}
	\includegraphics[width=\textwidth]{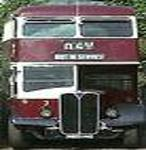}
	\end{subfigure}
	\begin{subfigure}{0.12\textwidth}
	\includegraphics[width=\textwidth]{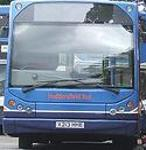}
	\end{subfigure}
	\begin{subfigure}{0.12\textwidth}
	\includegraphics[width=\textwidth]{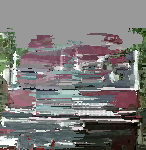}
	\end{subfigure}
	\begin{subfigure}{0.12\textwidth}
	\includegraphics[width=\textwidth]{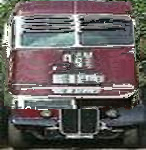}
	\end{subfigure}
	\begin{subfigure}{0.12\textwidth}
	\includegraphics[width=\textwidth]{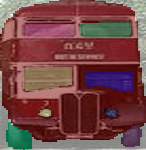}
	\end{subfigure}
	\begin{subfigure}{0.12\textwidth}
	\includegraphics[width=\textwidth]{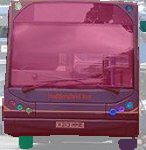}
	\end{subfigure}
	\begin{subfigure}{0.12\textwidth}
	\includegraphics[width=\textwidth]{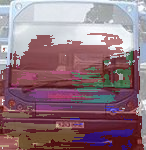}
	\end{subfigure}
	\begin{subfigure}{0.12\textwidth}
	\includegraphics[width=\textwidth]{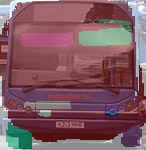}
	\end{subfigure}

	\begin{subfigure}{0.12\textwidth}
	\includegraphics[width=\textwidth]{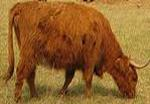}
	\end{subfigure}
	\begin{subfigure}{0.12\textwidth}
	\includegraphics[width=\textwidth]{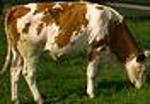}
	\end{subfigure}
	\begin{subfigure}{0.12\textwidth}
	\includegraphics[width=\textwidth]{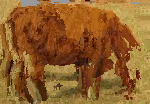}
	\end{subfigure}
	\begin{subfigure}{0.12\textwidth}
	\includegraphics[width=\textwidth]{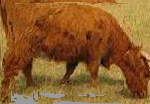}
	\end{subfigure}
	\begin{subfigure}{0.12\textwidth}
	\includegraphics[width=\textwidth]{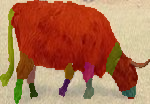}
	\end{subfigure}
	\begin{subfigure}{0.12\textwidth}
	\includegraphics[width=\textwidth]{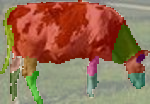}
	\end{subfigure}
	\begin{subfigure}{0.12\textwidth}
	\includegraphics[width=\textwidth]{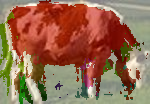}
	\end{subfigure}
	\begin{subfigure}{0.12\textwidth}
	\includegraphics[width=\textwidth]{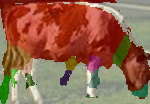}
	\end{subfigure}

	\begin{subfigure}{0.12\textwidth}
	\includegraphics[width=\textwidth]{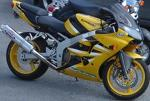}
	\end{subfigure}
	\begin{subfigure}{0.12\textwidth}
	\includegraphics[width=\textwidth]{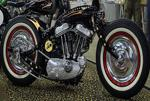}
	\end{subfigure}
	\begin{subfigure}{0.12\textwidth}
	\includegraphics[width=\textwidth]{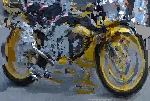}
	\end{subfigure}
	\begin{subfigure}{0.12\textwidth}
	\includegraphics[width=\textwidth]{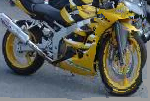}
	\end{subfigure}
	\begin{subfigure}{0.12\textwidth}
	\includegraphics[width=\textwidth]{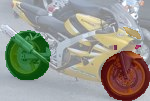}
	\end{subfigure}
	\begin{subfigure}{0.12\textwidth}
	\includegraphics[width=\textwidth]{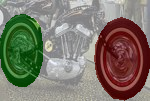}
	\end{subfigure}
	\begin{subfigure}{0.12\textwidth}
	\includegraphics[width=\textwidth]{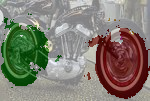}
	\end{subfigure}
	\begin{subfigure}{0.12\textwidth}
	\includegraphics[width=\textwidth]{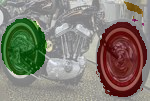}
	\end{subfigure}
	
	\hspace{0.3cm}
	\begin{subfigure}{0.08\textwidth}
	\includegraphics[width=\textwidth]{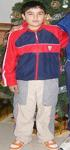}
	\end{subfigure}
	\hfill
	\begin{subfigure}{0.08\textwidth}
	\includegraphics[width=\textwidth]{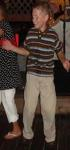}
	\end{subfigure}
	\hfill
	\begin{subfigure}{0.08\textwidth}
	\includegraphics[width=\textwidth]{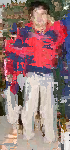}
	\end{subfigure}
	\hfill
	\begin{subfigure}{0.08\textwidth}
	\includegraphics[width=\textwidth]{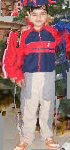}
	\end{subfigure}
	\hfill
	\begin{subfigure}{0.08\textwidth}
	\includegraphics[width=\textwidth]{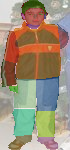}
	\end{subfigure}
	\hfill
	\begin{subfigure}{0.08\textwidth}
	\includegraphics[width=\textwidth]{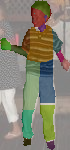}
	\end{subfigure}
	\hfill
	\begin{subfigure}{0.08\textwidth}
	\includegraphics[width=\textwidth]{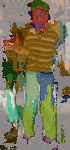}
	\end{subfigure}
	\hfill
	\begin{subfigure}{0.08\textwidth}
	\includegraphics[width=\textwidth]{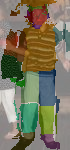}
	\end{subfigure}
	\hspace{0.3cm}

	\begin{subfigure}{0.12\textwidth}
	\includegraphics[width=\textwidth]{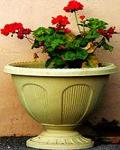}
	\end{subfigure}
	\begin{subfigure}{0.12\textwidth}
	\includegraphics[width=\textwidth]{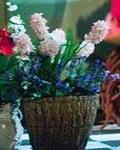}
	\end{subfigure}
	\begin{subfigure}{0.12\textwidth}
	\includegraphics[width=\textwidth]{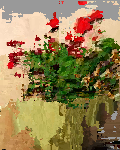}
	\end{subfigure}
	\begin{subfigure}{0.12\textwidth}
	\includegraphics[width=\textwidth]{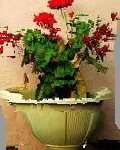}
	\end{subfigure}
	\begin{subfigure}{0.12\textwidth}
	\includegraphics[width=\textwidth]{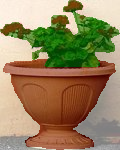}
	\end{subfigure}
	\begin{subfigure}{0.12\textwidth}
	\includegraphics[width=\textwidth]{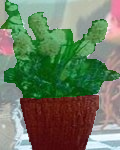}
	\end{subfigure}
	\begin{subfigure}{0.12\textwidth}
	\includegraphics[width=\textwidth]{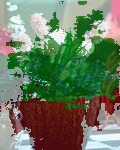}
	\end{subfigure}
	\begin{subfigure}{0.12\textwidth}
	\includegraphics[width=\textwidth]{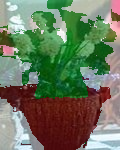}
	\end{subfigure}

	\begin{subfigure}{0.12\textwidth}
	\includegraphics[width=\textwidth]{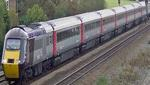}
	\end{subfigure}
	\begin{subfigure}{0.12\textwidth}
	\includegraphics[width=\textwidth]{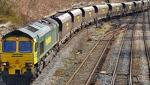}
	\end{subfigure}
	\begin{subfigure}{0.12\textwidth}
	\includegraphics[width=\textwidth]{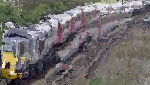}
	\end{subfigure}
	\begin{subfigure}{0.12\textwidth}
	\includegraphics[width=\textwidth]{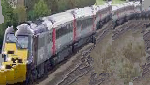}
	\end{subfigure}
	\begin{subfigure}{0.12\textwidth}
	\includegraphics[width=\textwidth]{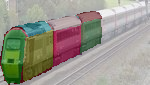}
	\end{subfigure}
	\begin{subfigure}{0.12\textwidth}
	\includegraphics[width=\textwidth]{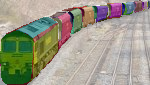}
	\end{subfigure}
	\begin{subfigure}{0.12\textwidth}
	\includegraphics[width=\textwidth]{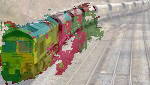}
	\end{subfigure}
	\begin{subfigure}{0.12\textwidth}
	\includegraphics[width=\textwidth]{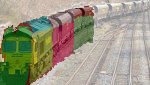}
	\end{subfigure}

	\begin{subfigure}{0.12\textwidth}
	\includegraphics[width=\textwidth]{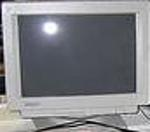}
	\caption{\footnotesize{Source image.}}
	\end{subfigure}
	\begin{subfigure}{0.12\textwidth}
	\includegraphics[width=\textwidth]{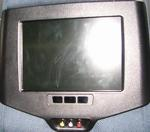}
	\caption{\footnotesize{Target image.}}
	\end{subfigure}
	\begin{subfigure}{0.12\textwidth}
	\includegraphics[width=\textwidth]{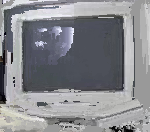}
	\caption{\footnotesize{DSP.}}
	\end{subfigure}
	\begin{subfigure}{0.12\textwidth}
	\includegraphics[width=\textwidth]{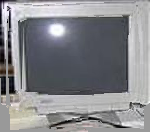}
	\caption{\footnotesize{Proposal Flow.}}
	\end{subfigure}
	\begin{subfigure}{0.12\textwidth}
	\includegraphics[width=\textwidth]{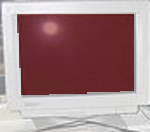}
	\caption{\footnotesize{Source mask.}}
	\end{subfigure}
	\begin{subfigure}{0.12\textwidth}
	\includegraphics[width=\textwidth]{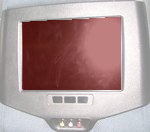}
	\caption{\footnotesize{Target mask.}}
	\end{subfigure}
	\begin{subfigure}{0.12\textwidth}
	\includegraphics[width=\textwidth]{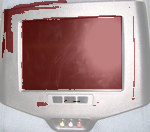}
	\caption{\footnotesize{DSP.}}
	\end{subfigure}
	\begin{subfigure}{0.12\textwidth}
	\includegraphics[width=\textwidth]{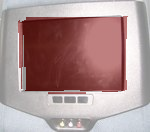}
	\caption{\footnotesize{Proposal Flow.}}
	\end{subfigure}
	\vfill
	\vspace{-0.2cm}
\caption{Examples of dense flow field on PASCAL parts. (a-b) Source images are warped to the target images using the dense correspondences estimated by (c) DSP~\cite{kim2013deformable} and (d) Proposal Flow w/ LOM, SS and HOG. (e-f) Similarly, annotated part segments for the source images are warped to the target images using the dense correspondences computed by (g) DSP and (h) Proposal Flow~w/~LOM, SS and HOG. (Best viewed in color.)}
\label{fig:voc_correspondence}	
\end{figure*}

\vspace{-0.5cm}
\paragraph{Matching results on Caltech-101.}
We also evaluate our approach on the Caltech-101
dataset~\cite{fei2006one}. Following the experimental protocol
in~\cite{kim2013deformable}, we randomly select 15 pairs of images for
each object class, and evaluate matching accuracy with three metrics:~Label transfer accuracy (LT-ACC)~\cite{liu2011nonparametric}, the IoU
metric, and the localization error (LOC-ERR) of corresponding pixel
positions. For LT-ACC, we transfer the class label of one image to the
other using dense correspondences, and count the number of correctly
labeled pixels. Similarly, the IoU score is measured between the
transferred label and ground truth.  Table \ref{tb:caltech} compares
quantitatively the matching accuracy of proposal flow to the state of
the art. It shows that proposal flow using LOM outperforms other
approaches, especially for the IoU score and the LOC-ERR of dense
correspondences. Note that compared to LT-ACC, these metrics evaluate
the matching quality for the foreground object, separate from 
irrelevant scene clutter. Our results verify that proposal flow
focuses on regions containing objects rather than scene clutter
and distracting details, enabling robust image matching against
outliers. %
\vspace{-0.5cm}
\paragraph{Matching results on PASCAL parts.}
We use the dataset provided by~\cite{zhou2015flowweb} where the images
are sampled from the PASCAL part dataset~\cite{chen2014detect}. We first measure part
matching accuracy using human-annotated part segments. For this
experiment, we measure the weighted IoU score between transferred
segments and ground truths, with weights determined by the pixel area
of each part (Table \ref{tb:voc}). To evaluate alignment accuracy, we
measure the PCK metric ($\alpha=0.05$) using keypoint annotations for
the 12 rigid PASCAL classes~\cite{xiang2014beyond}
(Table~\ref{tb:voc}). We use the same set of images as in the part
matching experiment. Proposal flow has an advantage over existing approaches on images that
contain cluttering elements (e.g., background, instance-specific texture, occlusion), but in this dataset~\cite{zhou2015flowweb}, such elements are confined
to only a small portion of the images, compared to the PF and the Caltech-101~\cite{fei2006one} datasets. This may be a reason that, for the PCK metric, our approach with
SS~\cite{uijlings2013selective} gives similar results to other methods. While FlowWeb~\cite{zhou2015flowweb} gives better results than ours, it relies on a cyclic constraint across multiple images (at least, three images). Thus, directly comparing our pairwise matching to FlowWeb is probably not fair. FlowWeb uses the output of DSP~\cite{kim2013deformable} as initial correspondences, and refines them with the cyclic constraint. Since our method clearly outperforms DSP, using FlowWeb as a post processing would likely increase performance. Figure~\ref{fig:voc_correspondence} visualize the part matching results.

 For more examples and qualitative results, see our project webpage.

\vspace{-0.3cm}
\section{Discussion}
\vspace{-0.2cm}
We have presented a robust region-based semantic flow method, called proposal
flow, and showed that it can effectively be mapped onto pixel-wise
dense correspondences. We have also introduced the PF 
dataset for semantic flow, and shown that it provides a reasonable benchmark for semantic
flow evaluation without extremely expensive manual annotation of full ground truth. 
Our benchmark can be used to evaluate region-based semantic flow methods and even 
pixel-based ones, and experiments with the PF dataset  
demonstrate that proposal flow substantially outperforms
existing semantic flow methods. Experiments
with Caltech and the VOC parts datasets further validate these results.

\vspace{-0.4cm}
\paragraph{Acknowledgments.}
This work was supported by the ERC grants VideoWorld and Allegro, and the Institut Universitaire de France.

{\small
\bibliographystyle{ieee}
\bibliography{proposal_flow}
}
\end{document}